\pdfoutput=1

\documentclass[11pt]{article}

\usepackage[final]{acl}

\usepackage{times}
\usepackage{latexsym}

\usepackage[T1]{fontenc}

\usepackage[utf8]{inputenc}

\usepackage{microtype}

\usepackage{inconsolata}

\usepackage{graphicx}

\usepackage{hyperref}
\usepackage{url}
\usepackage{xspace}
\usepackage{booktabs}
\usepackage{wrapfig}
\usepackage{float}
\usepackage{natbib}
\usepackage{graphicx}
\usepackage{xcolor}
\usepackage{amsmath}
\usepackage[utf8]{inputenc}
\usepackage[linesnumbered,ruled,vlined]{algorithm2e}
\usepackage{spverbatim}
\usepackage{hyperref}
\usepackage{soul}
\usepackage{tablefootnote}
\usepackage{multirow}
\usepackage[textsize=scriptsize]{todonotes}
\usepackage{hyperref}
\usepackage{colortbl} 
\usepackage{xcolor} 
\usepackage{booktabs}
\usepackage{amssymb}

\definecolor{gray}{rgb}{0.95, 0.95, 0.95} 

\newcommand{\postmark}{\textsc{PostMark}\xspace}
\newcommand{\emb}{\textsc{Embedder}\xspace}
\newcommand{\sectable}{\textsc{SecTable}\xspace}
\newcommand{\ins}{\textsc{Inserter}\xspace}

\definecolor{orange}{rgb}{0.85, 0.33, 0.2}

\newcommand{\opt}{\textsc{OPT-1.3B}\xspace}
\newcommand{\llama}{\textsc{Llama-3-8B}\xspace}
\newcommand{\llamainst}{\textsc{Llama-3-8B-Inst}\xspace}
\newcommand{\llamalinst}{\textsc{Llama-3-70B-Inst}\xspace}
\newcommand{\mistral}{\textsc{Mistral-7B-Inst}\xspace}
\newcommand{\gpt}{\textsc{GPT-4}\xspace}
\newcommand{\gptturbo}{\textsc{GPT-4-Turbo}\xspace}
\newcommand{\gpto}{\textsc{GPT-4o}\xspace}
\newcommand{\gpttturbo}{\textsc{GPT-3.5-Turbo}\xspace}
\newcommand{\openai}{\textsc{text-embedding-3-large}\xspace}
\newcommand{\nomic}{\textsc{nomic-embed}\xspace}
\newcommand{\paragram}{\textsc{paragram}\xspace}

\title{\postmark: A Robust Blackbox Watermark for Large Language Models}

\newcommand{\postbox}[0]{\includegraphics[width=.025\textwidth]{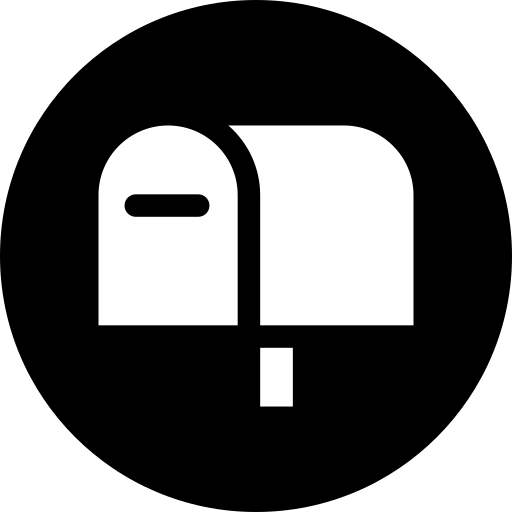}}
\newcommand{\stamp}[0]{\includegraphics[width=.025\textwidth]{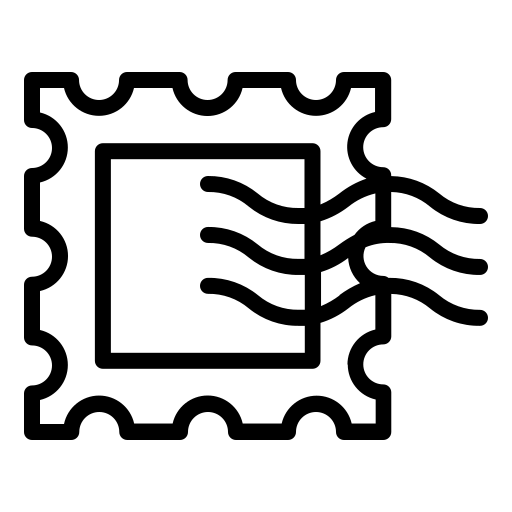}}

\author{\bf Yapei Chang\textsuperscript{\stamp} \quad Kalpesh Krishna\textsuperscript{\postbox} \\ \bf Amir Houmansadr\textsuperscript{\stamp} \quad John Wieting\textsuperscript{\postbox} \quad Mohit Iyyer\textsuperscript{\stamp} \\\\ \textsuperscript{\stamp}University of Massachusetts Amherst, \textsuperscript{\postbox}Google \\
\texttt{\{yapeichang,amir,miyyer\}@cs.umass.edu} \\
\texttt{\{kalpeshk,jwieting\}@google.com}\vspace{0.2cm}}

\begin{document}
\maketitle

\begin{abstract}
The most effective techniques to detect LLM-generated text rely on inserting a detectable signature---or \emph{watermark}---during the model's decoding process. Most existing watermarking methods require access to the underlying LLM's logits, which LLM API providers are loath to share due to fears of model distillation. As such, these watermarks must be implemented independently by each LLM provider. In this paper, we develop \postmark, a modular post-hoc watermarking procedure in which an input-dependent set of words (determined via a semantic embedding) is inserted into the text \emph{after} the decoding process has completed. Critically, \postmark\ does not require logit access, which means it can be implemented by a third party. We also show that \postmark\ is more robust to paraphrasing attacks than existing watermarking methods: our experiments cover eight baseline algorithms, five base LLMs, and three datasets. Finally, we evaluate the impact of \postmark on text quality using both automated and human assessments, highlighting the trade-off between quality and robustness to paraphrasing.
We release our code, outputs, and annotations at \href{https://github.com/lilakk/PostMark}{\path{https://github.com/lilakk/PostMark}}.
\end{abstract}

\section{Introduction}\label{sec:intro}

Large language models (LLMs) are increasingly being deployed for malicious applications such as fake content generation. The consequences of such applications for the web as a whole are dire: modern LLMs are known to hallucinate~\citep{xu2024hallucination}, and their outputs may contain biases and artifacts that are a product of their training data~\citep{navigli2023biases}. If the web is flooded with millions of LLM-generated articles, how can we trust the veracity of the content we are reading? Additionally, do we want to train LLMs of the future on text generated by LLMs of the present~\citep{shumailov2023curse}?

To combat this emerging problem, researchers have developed several \emph{LLM-generated text detection} techniques that leverage watermarking~\citep{aaronson2022, pmlr-v202-kirchenbauer23a}, outlier detection~\citep{mitchell2023detectgpt}, trained classifiers~\citep{2023gptzero}, or retrieval-based methods~\citep{krishna2023paraphrasing}. Among these, watermarking methods that embed detectable signatures into model outputs tend to be the most effective and robust~\citep{krishna2023paraphrasing}. However, most watermarking algorithms require access to the logits of the underlying LLM, which means that they can only be implemented by individual LLM API providers such as OpenAI or Google~\citep{yang2023blackbox}. Furthermore, while these methods are able to achieve high detection rates with minimal false positives, their effectiveness goes down when the LLM-generated text is modified through paraphrasing, translation, or cropping~\citep{krishna2023paraphrasing, he2024translation, kirchenbauer2024on}.

\begin{figure*}[t!]
\centering
\includegraphics[scale=0.175]{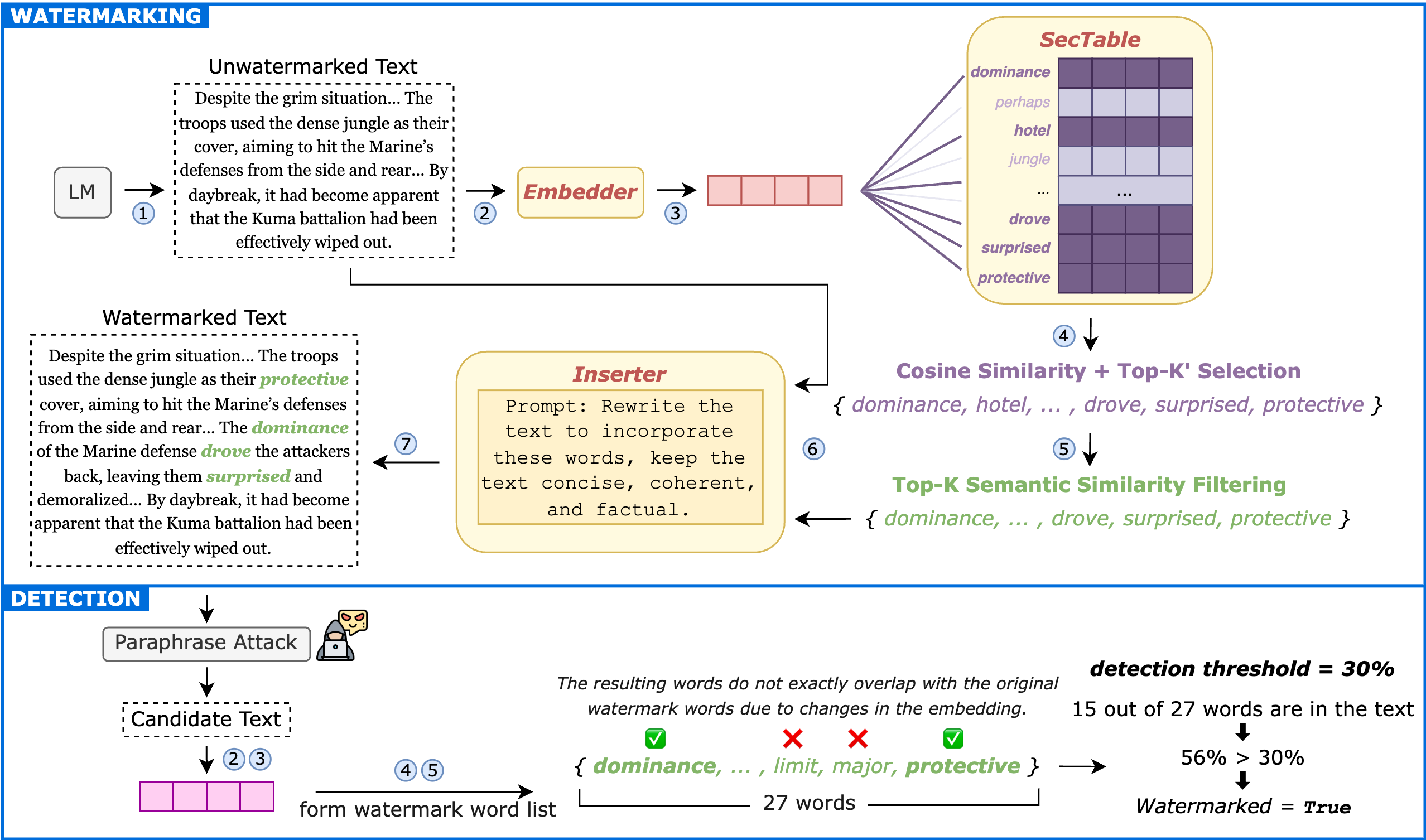}
\caption{The \postmark watermarking and detection procedure. Given some unwatermarked input text, we generate its embedding using the \emb and compute its cosine similarity with all word embeddings in the \sectable, performing top-k selection and additional semantic similarity filtering to choose a list of words. Then, we instruct the \ins to watermark the text by rewriting it to incorporate all selected words. During detection, we similarly obtain a watermark word list and check how many of these words are present in the input text.}
\label{fig:postmark}
\vspace{-0.15in}
\end{figure*}

In this work, we develop \postmark, a watermarking approach with relatively high detection rates even in the presence of paraphrasing attacks. \postmark is a \textbf{post}-hoc water\textbf{mark} that given some model-generated text, finds words conditioned on the semantics of the text using an embedding model, then calls a separate instruction-following LLM to insert these words into the text without appreciably modifying its meaning. Unlike prior methods, \postmark only requires access to the outputs of the underlying LLM (i.e., no logits).

Overall, our contributions are threefold: \textbf{1.} We propose \postmark, a novel post-hoc watermarking method that can be applied by third-party entities to outputs from an API provider like OpenAI. \textbf{2.} We conduct extensive experiments across eight baseline algorithms, five base LLMs, and three datasets, showing that \postmark offers superior robustness to paraphrasing attacks compared to existing methods. \textbf{3.} We verify through a human evaluation that the words inserted by \postmark during watermarking cannot be reliably detected by humans. We also conduct comprehensive quality evaluations encompassing \textit{coherence}, \textit{relevance}, and \textit{interestingness} for various watermarking methods. Notably, we also assess \textit{factuality}, an aspect that has not been evaluated in prior work. Our findings reveal that relatively robust watermarks all negatively affect factuality.

\section{\postmark: a post-hoc watermark}\label{sec:postmark}

Most existing watermarking algorithms embed the watermark during the LLM's decoding process. For example, the watermark of~\citet[][KGW]{pmlr-v202-kirchenbauer23a} partitions an LLM's vocabulary into two lists (a \emph{green} list and a \emph{red} list) at each decoding timestep based on a hash of the previous word, and then upweights the green tokens such that they are more likely to be sampled than red tokens.
These watermarks have several issues: (1) they require access to the LLM's logits; (2) because they rely on modifications to the next-token probability distribution, their effectiveness diminishes on LLMs that produce lower-entropy distributions, such as those that have undergone RLHF~\citep{bai2022training}; and (3) they show limited robustness to paraphrasing attacks as demonstrated by our results in~\autoref{sec:exp-results} and supported by findings from prior work~\citep{krishna2023paraphrasing, sadasivan2024aigenerated}.

In response, we develop \postmark, a watermarking method that does not require logit access, maintains high detection rates on low-entropy models and tasks, and exhibits improved robustness to paraphrasing attacks. Unlike existing watermarks, \postmark\ requires  access to just the text generated by the underlying LLM, not the next-token distributions. The rest of this section fully specifies \postmark's operation.

\paragraph{Intuition and terminology:} At a high level, \postmark is based on the intuition that a text's semantics should not drastically change after watermarking or paraphrasing. Thus, we can condition our watermark on a semantic embedding of the input text that ideally changes only minimally when paraphrasing is applied. To make this work, we rely on three modules: an 
embedding model \textbf{\emb}, a secret word embedding table \textbf{\sectable}, and an insertion model \textbf{\ins} implemented via an instruction-following LLM.

~\autoref{fig:postmark} illustrates \postmark's watermarking and detection pipelines. First, we generate the embedding of an input text using the \emb. We then compute the cosine similarity between this embedding and all of the word embeddings in \sectable, performing top-$k$ selection and filtering to form a watermark word list. Next, we prompt \ins to smoothly incorporating the selected words into the input to create the watermarked text. During detection, we follow similar steps to obtain a word list, and check how many of the words are present in the input text.

\paragraph{Embedding model \emb:} The \emb needs to be capable of projecting both words and documents into a high-dimensional latent space. In our main experiments, we use OpenAI's \openai~\citep{2024openaiemb}, a powerful model that demonstrates strong performance on the MTEB benchmark~\citep{muennighoff-etal-2023-mteb}. However, any embedding model can be used here. In~\autoref{sec:exp-results}, we also experiment with \nomic~\citep{nussbaum2024nomic}, an open-source model.

\paragraph{Secret word embedding table \sectable:} The core idea behind \postmark is to use an LLM to insert a list of watermark words into the input text without appreciably modifying the quality or meaning of the text, where the words in the list are selected by computing the cosine similarity between the text embedding and a word embedding table \sectable. The construction of \sectable involves two main steps, which we detail below:

\vspace{2pt}
\noindent\emph{\textcolor{orange}{\textbf{> Step 1.} Choosing a vocabulary $\mathbb{V}$:}} To decide which words to include in \sectable, we use the WikiText-103 corpus~\citep{merity2017pointer} as our base vocabulary. To avoid inserting arbitrary words that make little sense, we remove all function words, proper nouns, and infrequent rare words. This refined set forms our final vocabulary, $\mathbb{V}$. We provide more details on this filtering process in~\S\ref{appendix-sectable}.

\noindent\emph{\textcolor{orange}{\textbf{> Step 2.} Mapping words in $\mathbb{V}$ to embeddings:}} To make it difficult for attackers to recover our embedding table, we construct \sectable\ by randomly assigning each word in the vocabulary to an embedding produced by \emb; the resulting mapping acts as a cryptographic key.\footnote{We could also just use  \emb's word embeddings as \sectable\ directly. However, this can easily be recovered by an attacker, and our experiments show that it also reduces \postmark's effectiveness due to many words already being present in the input text even before insertion.} More specifically, we generate a set of embeddings $\mathbb{D}$ for a collection of random documents using \emb\ and then randomly map each word in $\mathbb{V}$ to a unique document embedding in $\mathbb{D}$ to produce \sectable.\footnote{The selection of these documents is flexible. In our experiments, we randomly sample 250-word snippets from the RedPajama dataset's English split~\citep{together2023redpajama}.}

\paragraph{Insertion model \ins:} The \ins needs to have instruction-following capabilities, and its purpose is to rewrite the input text to incorporate words from the watermark word list. We use \gpto~\citep{openai-gpt4o} as the \ins in our main experiments, and later show in~\autoref{sec:exp-results} that open-source models like \llamalinst~\citep{llama3modelcard} also show promising performance.

\subsection{Inserting the watermark}

\noindent\emph{\textcolor{orange}{\textbf{> Step 1.} Deciding how many words to insert:}} How many words should we insert into a given text? We define a hyperparameter called the \textbf{insertion ratio} $r$ that determines this number. The insertion ratio represents the percentage of the input text's word count: for example, if $r = 10\%$ and the input text has $50$ words, we will insert 5 words.

\noindent\emph{\textcolor{orange}{\textbf{> Step 2.} Obtaining a watermark word list:}} Suppose that the watermark list should contain $k$ words. To create the watermark word list given the input text, we first compute the input's embedding $e_t = \emb (t)$. Next, we compute \text{CosineSimilarity}$(e_t, \sectable)$ and select the top $k'$ most similar words, then perform semantic similarity filtering to obtain the final $k$ words.\footnote{Due to the random nature of the word-to-embedding mapping of $T$, the top $k'$ words might include highly irrelevant words (e.g., ``hotel'' in ~\autoref{fig:postmark}). Thus, we refine the top-$k'$ list by selecting the top $k$ words whose \textit{actual} embeddings (as determined by \emb) are most similar to $e_t$.} We present an analysis on how frequently a word is chosen as an watermark word in~\S\ref{appendix-sectable}.

\noindent\emph{\textcolor{orange}{\textbf{> Step 3.} Inserting words into the text:}} To watermark the text, we instruct \ins to rewrite it via zero-shot prompting, incorporating words in the watermark word list while keeping the rewritten text coherent, factual, and concise.\footnote{In practice, we find that dividing a long word list into sublists of 10 words each and then iteratively asking the \ins to incorporate each sublist ensures a high insertion success rate. This may not be necessary if the \ins has better instruction-following capabilities.} The prompt can be found in~\S\ref{appendix-insertion-prompt}. 

\subsection{Detecting the watermark} During detection, given some text, the goal is to find out if the text contains a watermark. Similar to the watermarking procedure, we embed the candidate text using \emb, form a word list, and then check how many words in the list are present in the text by computing a presence score $p$:

\[
p = \frac{ \left| \left\{ w \in \text{list} \ \text{s.t.} \ \exists w' \in \text{text}, \ \text{sim}(w', w) \geq 0.7 \right\} \right| }{|\text{list}|}
\]

A word $w$ is marked present in the text if there is any other word $w'$ with an embedding cosine similarity greater than a threshold that we set to 0.7. We choose this method over exact match to ensure additional robustness against paraphrasing.\footnote{We use the \paragram\ word embedding model developed by~\citet{wieting-etal-2015-paraphrase} to compute cosine similarity for this step. This model is chosen for its superior performance in assigning high similarity scores to close synonyms and low scores to unrelated words, more details in~\S\ref{appendix-cos-sim-detection}.} 
If $p$ is larger than a certain threshold, it is likely that the text has been watermarked. As later discussed in~\autoref{sec:exp-setup}, the primary metric we use to measure detection accuracy is the true positive rate at a fixed 1\% false positive rate. We thus set the threshold to ensure a 1\% FPR, same as what we do for all baselines in our main experiments.

\section{Experiments}\label{sec:exp}

In this section, through extensive experiments on three datasets and five language models, we demonstrate that \postmark consistently outperforms both logit-free and logit-based methods in terms of robustness to paraphrasing attacks, especially on low-entropy models that have undergone RLHF alignment. Furthermore, we showcase \postmark's modular design by testing an open-source variant, which achieves promising results.

\subsection{Experimental setup} \label{sec:exp-setup}

\paragraph{Baselines:} We compare \postmark against 8 baseline algorithms, more detailed descriptions can be found in~\S\ref{appendix-baseline}. (1) \textbf{KGW}~\citep{pmlr-v202-kirchenbauer23a}: Partitions the vocabulary into ``green'' and ``red'' lists based on the previous token, then boosts the probability of green tokens during generation. (2) \textbf{Unigram}~\citep{zhao2023provable}: A more robust variant of KGW that uses a fixed partition for all generations. (3) \textbf{EXP}~\citep{aaronson2022}: Uses exponential sampling to bias token selection with a pseudo-random sequence. (4) \textbf{EXP-Edit}~\citep{kuditipudi2024robust}: A variant of EXP that uses edit distance during detection. (5) \textbf{SemStamp}~\citep{hou-etal-2023-semstamp}: A sentence-level algorithm that partitions the sentence semantic space. (6) \textbf{k-SemStamp}~\citep{hou2024ksemstamp}: Improves SemStamp by using k-means clustering to partition the semantic space. (7) \textbf{SIR}~\citep{liu2024a-sir}: Generates watermark logits from the semantic embeddings of preceding tokens then adds them to the model's logits. (8) \textbf{Blackbox}~\citep{yang2023blackbox}: This method, like ours, works in a blackbox setting where only model outputs are visible. It substitutes words representing bit-0 in a binary encoding scheme with synonyms representing bit-1.

\paragraph{Hyperparameters:} The key hyperparameter for \postmark is the insertion ratio $r$, which controls how many words are inserted during the watermarking process. We set $r$ to 12\% as preliminary experiments suggest that this value strikes a good balance between quality and robustness. \autoref{sec:quality-auto-eval} explores different \postmark configurations that vary $r$. In all following discussion and tables, we refer to these configurations with the naming convention ``\textsc{PostMark}@$r$''. We carefully tune all baselines' hyperparameters to maximize their robustness to paraphrasing; more details  in~\S\ref{appendix-baseline}.

\paragraph{Base models:} Our experiments involve five generative models: \llama~\citep{llama3modelcard}, \llamainst~\citep{llama3modelcard}, \mistral~\cite{jiang2023mistral}, \gpt~\citep{openai2024gpt4}, and \opt~\citep{zhang2022opt}. Among these, \llamainst, \mistral, and \gpt have been aligned with human preferences. For details on model checkpoints and generation length, see~\S\ref{appendix-base-models}. We do not run \opt ourselves but directly use its unwatermarked outputs provided by~\citet{hou2024ksemstamp}. Due to difficulties in running SemStamp, k-SemStamp, and SIR,\footnote{Their code is available but not runnable yet. We look forward to running these methods ourselves once the issues are resolved.} we apply \postmark directly to these outputs and compare our results with the published numbers in~\citet{hou2024ksemstamp}.

\paragraph{Datasets:} Our main experiments use three datasets: (1) \textbf{OpenGen}, a dataset collected by~\citet{krishna2023paraphrasing} designed for open-ended generation that consists of two-sentence chunks sampled from the validation set of WikiText-103; (2) \textbf{LFQA}, a dataset collected by~\citet{krishna2023paraphrasing} for long-form question answering that contains questions sampled from the \texttt{r/explainlikeimfive} subreddit that span multiple domains; and (3) \textbf{RealNews}~\citep{raffel2020c4}, a subset of the C4 dataset that includes news articles gathered from a wide range of reliable news websites.

\paragraph{Paraphrasing attack setup:} Following prior work~\citep{hou-etal-2023-semstamp, hou2024ksemstamp, kirchenbauer2024on, liu2024a-sir}, we use \gpttturbo as our paraphraser. We use a sentence-level paraphrasing approach where the model iterates through each sentence of the input text, using all preceding context to paraphrase the current sentence. See~\S\ref{appendix-para} for more details on this setup.

\paragraph{Metric for measuring detection performance:} In addition to the true positive rate, a low false positive rate is critical for LLM-generated detection. Thus, following prior detection work~\citep{krishna2023paraphrasing, zhao2023provable, hou-etal-2023-semstamp, hou2024ksemstamp, liu2024a-sir}, we use \textbf{TPR at 1\% FPR} as our primary metric.

\begin{table*}[ht]
\renewcommand{\arraystretch}{1.2}
\centering
\scriptsize
\resizebox{\textwidth}{!}{
\begin{tabular}{llcccccccccc}
\toprule
Metric $\rightarrow$ & & & \multicolumn{9}{c}{ TPR at 1\% FPR (\textcolor{purple}{Before Paraphrasing} / \textcolor{teal}{After Paraphrasing}) } \\
\cline{4-12}
Model $\downarrow$ & Dataset $\downarrow$ & Avg Entropy $\downarrow$ & \textsc{PostMark}@12 & Blackbox & KGW & Unigram & EXP & EXP-Edit & SIR & SemStamp & k-SemStamp \\
\midrule 
\llama & OpenGen & 3.6 & \textcolor{purple}{99.7} / \textcolor{teal}{63.5} & \textcolor{purple}{81.2} / \textcolor{teal}{2.2} & \textcolor{purple}{100} / \textcolor{teal}{74.8} & \textcolor{purple}{99.8} / \textcolor{teal}{\textbf{93.4}} & \textcolor{purple}{99.8} / \textcolor{teal}{36.6} & \textcolor{purple}{97.3} / \textcolor{teal}{73.3} & - & - & - \\
 & LFQA & 3.5 & \textcolor{purple}{97.8} / \textcolor{teal}{72.5} & \textcolor{purple}{82.8} / \textcolor{teal}{1.6} & \textcolor{purple}{99.8} / \textcolor{teal}{25.6} & \textcolor{purple}{99.8} / \textcolor{teal}{\textbf{79.6}} & \textcolor{purple}{99.8} / \textcolor{teal}{12.4} & \textcolor{purple}{83} / \textcolor{teal}{41} & - & - & - \\
\midrule 
\llamainst & OpenGen & 1.6 & \textcolor{purple}{99.4} / \textcolor{teal}{\textbf{46.4}} & \textcolor{purple}{91.8} / \textcolor{teal}{1} & \textcolor{purple}{98.2} / \textcolor{teal}{21.6} & \textcolor{purple}{99.6} / \textcolor{teal}{41.4} & \textcolor{purple}{99.6} / \textcolor{teal}{4.8} & \textcolor{purple}{47.8} / \textcolor{teal}{2.2} & - & - & - \\
 & LFQA & 1.3 & \textcolor{purple}{96} / \textcolor{teal}{\textbf{65.7}} & \textcolor{purple}{86.2} / \textcolor{teal}{3} & \textcolor{purple}{85.8} / \textcolor{teal}{19} & \textcolor{purple}{98.6} / \textcolor{teal}{31.8} & \textcolor{purple}{98.4} / \textcolor{teal}{0.6} & \textcolor{purple}{21.1} / \textcolor{teal}{0.6} & - & - & - \\
\midrule 
\mistral & OpenGen & 1.4 & \textcolor{purple}{99.2} / \textcolor{teal}{\textbf{69.2}} & \textcolor{purple}{98.4} / \textcolor{teal}{0.4} & \textcolor{purple}{100} / \textcolor{teal}{16} & \textcolor{purple}{99.8} / \textcolor{teal}{56} & \textcolor{purple}{99.4} / \textcolor{teal}{5} & \textcolor{purple}{33} / \textcolor{teal}{1.5} & - & - & - \\
 & LFQA & 1.1 & \textcolor{purple}{99.6} / \textcolor{teal}{\textbf{56.4}} & \textcolor{purple}{89.8} / \textcolor{teal}{0.4} & \textcolor{purple}{99.4} / \textcolor{teal}{23.6} & \textcolor{purple}{97.2} / \textcolor{teal}{41.2} & \textcolor{purple}{97.4} / \textcolor{teal}{0.8} & \textcolor{purple}{20.1} / \textcolor{teal}{2.1} & - & - & - \\
\midrule 
\gpt & OpenGen & - & \textcolor{purple}{99.4} / \textcolor{teal}{\textbf{59.4}} & \textcolor{purple}{99.4} / \textcolor{teal}{1.4} & - & - & - & - & - & - & - \\
 & LFQA & - & \textcolor{purple}{99.4} / \textcolor{teal}{\textbf{65}} & \textcolor{purple}{99.2} / \textcolor{teal}{0.4} & - & - & - & - & - & - & - \\
\midrule 
\opt & RealNews & 3.6 & \textcolor{purple}{98.2} / \textcolor{teal}{67.2} & \textcolor{purple}{1.2} / \textcolor{teal}{0} & \textcolor{purple}{99.2} / \textcolor{teal}{40.8} & \textcolor{purple}{98.8} / \textcolor{teal}{77.2} & \textcolor{purple}{99.4} / \textcolor{teal}{\textbf{80.7}} & \textcolor{purple}{69.6} / \textcolor{teal}{46.7} & \textcolor{purple}{99.4} / \textcolor{teal}{24.7} & \textcolor{purple}{93.9} / \textcolor{teal}{33.9} & \textcolor{purple}{98.1} / \textcolor{teal}{55.5} \\
\bottomrule 
\end{tabular}
}
\caption{Comparison of \textsc{PostMark} and baselines. All numbers are computed over 500 generations. Each entry shows the TPR at 1\% FPR \textcolor{purple}{before paraphrasing} and \textcolor{teal}{after paraphrasing}. The ``Avg Entropy'' column shows the average token-level entropy (in bits) of each model on each dataset.}
\label{tab:tpr-all}
\vspace{-0.15in}
\end{table*}

\subsection{Results} \label{sec:exp-results}

We present our main experimental results on robustness to paraphrasing attacks in~\autoref{tab:tpr-all}, and discuss our main findings below. Runtime analysis and API cost estimates can be found in~\S\ref{appendix-runtime}.

\paragraph{\postmark is an effective and robust watermark.} \postmark consistently achieves a high TPR \textcolor{purple}{before paraphrasing} ($>90\%$), outperforming baselines like Blackbox, KGW, and EXP-Edit. Additionally, \postmark achieves higher TPR \textcolor{teal}{after paraphrasing} compared to other baselines, including Blackbox, the only other method that operates under the same logit-free condition. The only settings that \postmark\ is not the most robust model under paraphrasing is with \llama\ and \opt, where Unigram and EXP respectively exhibit more robustness. We note that Unigram is much more vulnerable to reverse-engineering than \postmark\ because it uses a fixed \textit{green}/\textit{red} list partition for all inputs, which can be exploited with repetition attacks.\footnote{For Unigram, detection works by comparing the number of \textit{green} tokens present in the input text to the expected count under the null hypothesis of no watermarking. The adversary can pick a word ``apple'' and submit a long repeating sequence of this word (e.g., ``apple apple apple...'') to the watermark detection service. If it says this sequence is watermarked, then ``apple'' must be in the green list.} Both Unigram and EXP's effectiveness diminish with low-entropy models. In~\S\ref{appendix-unigram-rep} and~\S\ref{appendix-exp-rep}, we also observe that both methods significantly degrade text quality, leading to excessive repetitions.

\paragraph{Logit-based baselines perform worse on low-entropy models and tasks, while \postmark stays relatively unaffected.} Results from~\autoref{tab:tpr-all} demonstrate that logit-based baselines (i.e., all baselines except Blackbox) generally perform worse on aligned models (\llamainst and \mistral) compared to the non-aligned \llama, and worse on LFQA than on OpenGen. This performance difference is consistent with findings from prior work \citep{kuditipudi2024robust} and can be attributed to the lower entropy of aligned models resulting from RLHF or instruction-tuning, as well as the inherently lower entropy of the LFQA task. The ``Avg Entropy'' column of~\autoref{tab:tpr-all} illustrates these entropy differences. In contrast, \postmark consistently outperforms all baselines in terms of robustness against paraphrasing attacks in these low-entropy scenarios.

\paragraph{Open-weight \postmark shows promise.} While our main experiments use \gpto as the \ins and OpenAI's \openai as the \emb, we show in~\autoref{tab:tpr-open-source} that an open-weight combination of \llamalinst and \nomic can also achieve promising robustness to paraphrasing attacks. The modular design of \postmark allows for flexible experimentation with various components. As each module's capabilities advance, \postmark's robustness will likewise improve.

\begin{table}[ht]
\centering
\small
\begin{tabular}{lc}
\toprule 
\textsc{PostMark}@12 Impl. & TPR at 1\% FPR \\
\midrule 
Closed & \textcolor{purple}{99.4} / \textcolor{teal}{59.4} \\
Open & \textcolor{purple}{100} / \textcolor{teal}{52.1} \\
\bottomrule
\end{tabular}
\caption{TPR at 1\% FPR \textcolor{purple}{before} and \textcolor{teal}{after paraphrasing}. The open-source implementation of \textsc{PostMark}@12 with \nomic as the \emb and \llamalinst as the \ins shows promising performance on OpenGen with \gpt as the base LLM.}
\label{tab:tpr-open-source}
\vspace{-0.15in}
\end{table}

\section{Impact of watermarking on text quality}\label{sec:quality}


\begin{table*}[ht]
\centering
\footnotesize
\begin{tabular}{p{0.1\textwidth} p{0.35\textwidth} p{0.47\textwidth}}
\toprule
\multicolumn{1}{c}{\textsc{Type}} & \multicolumn{1}{c}{\textsc{Before watermark}} & \multicolumn{1}{c}{\textsc{After watermark}} \\
\midrule
\multicolumn{3}{c}{\textbf{Rewriting existing content}} \\
\midrule
Rewording & \textit{Her decision to quit the opera, however, did not lessen the engulfing sadness which veiled her once radiant joy.} & \textit{Her decision to \textcolor{orange}{\textbf{resign} from} the opera, however, did not lessen the engulfing sadness which veiled her once radiant joy.} \\
\midrule
Clarification & \textit{Since the charges concerned violation of civil rights and not actual murder, the defendants received surprisingly light sentences, ranging from three to ten years.} & \textit{Since the charges concerned violation of civil rights and not actual murder, the defendants received surprisingly light sentences, ranging from three to ten years \textcolor{orange}{of \textbf{imprisonment}.}} \\
\midrule
\multicolumn{3}{c}{\textbf{Adding new content}} \\
\midrule
Metaphors & \textit{In fact, despite Mount Elbert's somewhat minimal precipitation, it displays a remarkable ability to sustain life.} & \textit{In fact, despite Mount Elbert's somewhat minimal precipitation, it displays a remarkable ability to sustain life, \textcolor{orange}{almost as if it wears an \textbf{armor} of resilience, \textbf{immune} to the challenges it faces.}} \\
\midrule
Interpretive claims & \textit{He swiftly plants timed explosives around the warehouse, ensuring to place a few on the largest weapon caches for maximum effect.} & \textit{He swiftly plants timed explosives around the warehouse, ensuring to place a few on the largest weapon caches for maximum effect. \textcolor{orange}{The \textbf{depth} of his planning was a testament to his expertise in \textbf{defense} tactics.}} \\
\midrule
New details & \textit{Headlam had the ability to foster a culture of discipline, camaraderie and respect among the airmen under his command, reflecting his firm belief in focused team effort and mutual support.} & \textit{Headlam had the ability to foster a culture of discipline, camaraderie and respect among the airmen under his command, reflecting his firm belief in focused team effort and mutual support. \textcolor{orange}{His attention to \textbf{detail} was evident in every aspect of the unit's operations.}} \\
\bottomrule
\end{tabular}
\caption{Example edits made by \postmark during the watermarking process. Changes are highlighted in \textcolor{orange}{orange}, and watermark words are in \textcolor{orange}{\textbf{bold}}.}
\label{tab:examples}
\vspace{-0.1in}
\end{table*}

\begin{table}[ht]
\centering
\small
\begin{tabular}{lc}
\toprule
Base LLM & SIM  \\
\midrule
\llama     & 94.2 \\
\llamainst & 94.8 \\
\mistral   & 94.6 \\
\gpt       & 95.3 \\
\bottomrule
\end{tabular}
\caption{Average cosine similarity between the embeddings of unwatermarked and \textsc{PostMark}@12 watermarked outputs on OpenGen. Embeddings are obtained using \openai. Numbers are averaged over 500 pairs.}
\label{tab:semantic}
\end{table}

\postmark modifies text during watermarking by inserting new words, which often results in longer watermarked text.
\footnote{A full table of length comparison is in~\S\ref{appendix-length}.}~\autoref{tab:examples} shows several common types of edits made by \postmark during watermarking.\footnote{Summarized based on a small-scale qualitative analysis.} Although edits adding new content are expected to hurt quality, this quality degradation is not unique to \postmark. Prior work has found that all watermarking methods negatively affect text quality to some extent~\citep{singh2023new}. For logit-based methods like KGW, quality degradation occurs because relevant words can be downweighted during decoding. While existing papers on watermarking often lack extensive quality evaluations, we conduct both automatic and human evaluations to assess the quality of watermarked text  (\textit{relevance}, \textit{coherence}, \textit{interestingness}, and \textit{factuality}) in this section.

\paragraph{Semantic meaning preservation} To check whether \postmark preserves the general semantic meaning of the original unwatermarked text, we compute the average cosine similarity between the embeddings unwatermarked and watermarked outputs in~\autoref{tab:semantic}, and find the similarity score to be consistently around 0.95.

\paragraph{Setting up quality evaluations:} Prior work on watermarking has predominantly used perplexity as a measure for text quality~\citep{pmlr-v202-kirchenbauer23a, zhao2023provable, yang2023blackbox, liu2024a-sir, hu2024unbiased, hou-etal-2023-semstamp, hou2024ksemstamp}. However, perplexity alone has been shown to be an unreliable indicator of quality~\citep{Wang2022PerplexityFP}. Some studies have explored alternative methods, such as LLM-based evaluations~\citep{singh2023new} and human assessments~\citep{kirchenbauer2024on}. Here, we evaluate the quality of watermarked text using automated and human evaluations, aiming to address four key questions:

\noindent\emph{\textcolor{orange}{\textbf{> Q1:}} How does \postmark compare to other baselines in terms of impact on text quality?}

\noindent\emph{\textcolor{orange}{\textbf{> Q2:}} What is the quality-robustness trade-off for \postmark?}

\noindent\emph{\textcolor{orange}{\textbf{> Q3:}} How often do humans think that \postmark watermarked texts are at least as good as their unwatermarked versions?}

\noindent\emph{\textcolor{orange}{\textbf{> Q4:}} Are words inserted by \postmark easily detectable by humans?}

\subsection{Automatic evaluation} \label{sec:quality-auto-eval}

In this section, we compare \postmark with other baselines regarding impact on quality (\textcolor{orange}{\emph{\textbf{Q1}}}) and address the quality-robustness trade-off of \postmark (\textcolor{orange}{\emph{\textbf{Q2}}}).

\paragraph{Pairwise preference evaluation setup:} We adopt the LLM-as-a-judge~\citep{zheng2023judging} setup to perform a pairwise comparison task. We choose \gptturbo as our judge as it is the high-ranked evaluator model on the Reward Bench leaderboard~\citep{lambert2024rewardbench}\footnote{The current leaderboard is hosted on \href{https://huggingface.co/spaces/allenai/reward-bench}{huggingface}. \gptturbo's high ranking indicates that it is a relatively robust and reliable LLM evaluator.} that we can easily access. Given 100 OpenGen prefixes and corresponding pairs of anonymized unwatermarked and watermarked responses, the model evaluates each pair and chooses which response it prefers, where ties are allowed. The model is instructed to consider the \textit{relevance}, \textit{coherence}, and the \textit{interestingness} of the responses when making a judgment. The full prompt can be found in~\S\ref{appendix-auto-eval}. Then, we compute the \textbf{soft win rate} of various baselines in~\autoref{tab:quality-comparison} and several \postmark\ configurations in ~\autoref{tab:quality-tradeoff}, which equals the number of ties plus the number of wins for the watermarked response.

\begin{table}[ht]
\centering
\vspace{5pt}
\small
\begin{tabular}{lc}
\toprule
 & FactScore\\
\midrule 
\llamainst & 40.2 \\
+ KGW & 37.8 \\
+ Unigram & 37.2 \\
+ \textsc{PostMark}@12 & 37.3 \\
+ \textsc{PostMark}@6 & 38.3 \\
\end{tabular}
\caption{FactScore evaluation results based on 100 generations with \llamainst as the base generator LLM. All four evaluated methods impact factuality negatively to some extent, with less robust methods causing a lesser negative impact.}
\label{tab:factscore}
\vspace{-0.15in}
\end{table}

\paragraph{Factuality evaluation setup:} To assess \textit{factuality}, an essential aspect not addressed in the previous pairwise comparisons \emph{or} previous watermarking research, we use FactScore~\citep{min-etal-2023-factscore}, an automatic metric that measures the percentage of atomic claims in an LLM-generated biography that are supported by Wikipedia. 
We generate biographies for the entities in the FactScore dataset and comparing the FactScores of the outputs before and after watermarking. Results are reported in~\autoref{tab:factscore}. Before watermarking, \llamainst achieves a score of 40.2. Running KGW, Unigram, \textsc{PostMark}@12, and \textsc{PostMark}@6 all results in slight reductions in FactScore. Overall, less robust methods (KGW and \textsc{PostMark}@6) have less negative impact on factuality.

\begin{table}[ht]
\renewcommand{\arraystretch}{1.2}
\centering
\tiny
\begin{tabular}{l@{\hspace{2.5pt}}c@{\hspace{2.5pt}}c@{\hspace{2.5pt}}c@{\hspace{2.5pt}}c@{\hspace{2.5pt}}c@{\hspace{2.5pt}}c@{\hspace{2.5pt}}c}
\toprule
Metric $\rightarrow$ & \multicolumn{6}{c}{ Soft Win Rate } \\
\cline{2-7}
Method $\rightarrow$ & KGW & Unigram & EXP & EXP-Edit & Blackbox & \textsc{PostMark}@12 \\ \hline
\llama     & 37 & 17 & 23 & 49 & 45 & \textbf{74} \\
\llamainst & 52 & 52 & 59 & 57 & 55 & \textbf{68} \\
\mistral   & 57 & 54 & 49 & 54 & 46 & \textbf{64} \\
\gpt       & - & - & - & - & 53 & \textbf{64} \\
\bottomrule
\end{tabular}
\caption{Soft win rates computed based on the pairwise comparison evaluation with \gptturbo as the judge, measured over 100 pairs of unwatermarked and watermarked OpenGen outputs from various LLMs (first column). \textsc{PostMark}@12 outperforms all baselines.}
\label{tab:quality-comparison}
\vspace{-0.15in}
\end{table}

\begin{table}[ht]
\centering
\small
\begin{tabular}{lcc}
\toprule Configuration & Soft Win Rate & TPR \textcolor{teal}{After Para.} \\
\midrule
\textsc{PostMark}@6 & 84 & \textcolor{teal}{20.8} \\
\textsc{PostMark}@8 & 79 & \textcolor{teal}{28.2} \\
\textsc{PostMark}@12 & 64 & \textcolor{teal}{59.4} \\
\textsc{PostMark}@15 & 67 & \textcolor{teal}{61.9} \\
\textsc{PostMark}@20 & 62 & \textcolor{teal}{82.8} \\
\textsc{PostMark}@30 & 55 & \textcolor{teal}{98} \\
\bottomrule 
\end{tabular}
\caption{Quality-robustness trade-off. All soft win rates are averaged over 100 pairs of unwatermarked and watermarked texts judged by \gptturbo. All paraphrased TPR numbers at 1\% FPR are computed over on 500 OpenGen instances.}
\label{tab:quality-tradeoff}
\vspace{-0.15in}
\end{table}

\paragraph{\textcolor{orange}{\emph{\textbf{> Q1:}}} \postmark does not affect quality as much as other baselines.} Results from \autoref{tab:quality-comparison} show that \postmark performs exceptionally well in pairwise comparisons across models. In contrast, despite Unigram's strong robustness to paraphrasing—sometimes even outperforming \postmark when tested on \llama—it has a significantly lower soft win rates, especially on \llama (17\%). This low score is likely due to frequent repetitions in Unigram outputs, as detailed in~\S\ref{appendix-unigram-rep}. Regarding factuality, KGW, Unigram, and \textsc{PostMark}@12 all show similar levels of negative impact as their FactScores are respectively 37.8, 37.2, and 37.3.

\paragraph{\textcolor{orange}{\emph{\textbf{> Q2:}}} Inserting more words enhances robustness but hurts quality, and vice versa.} We first use the pairwise comparison setup to evaluate the quality-robustness trade-off of \postmark with $r$ set to six different values: 6, 8, 12, 15, 20, and 30. Results in~\autoref{tab:quality-tradeoff} reveal a strong inverse correlation between quality and robustness, with a Pearson coefficient of -0.98. \textsc{FactScore}@6 also achieves a higher FactScore (38.3) than \textsc{FactScore}@12 (37.3). In practical applications, the choice of $r$ should be based on the desired balance between quality and robustness.

\subsection{Human evaluation} \label{sec:quality-human-eval}

While LLM-based evaluators serve as good proxies for human judgments in several cases~\citep{zheng2023judging}, their results should be interpreted with caution, as they can be biased to certain aspects of the text such as length~\citep{wang2023large} or overlap between the generator and the judge model~\citep{panickssery2024llm}. Thus, we hire two annotators fluent in English and conduct two human annotation studies detailed below, addressing \textcolor{orange}{\emph{\textbf{Q3}}} and \textcolor{orange}{\emph{\textbf{Q4}}}. More details on annotator qualifications, payment, and each annotation setup can be found in~\S\ref{appendix-human-eval}.

\begin{figure}[htbp]
\centering
\includegraphics[scale=0.23]{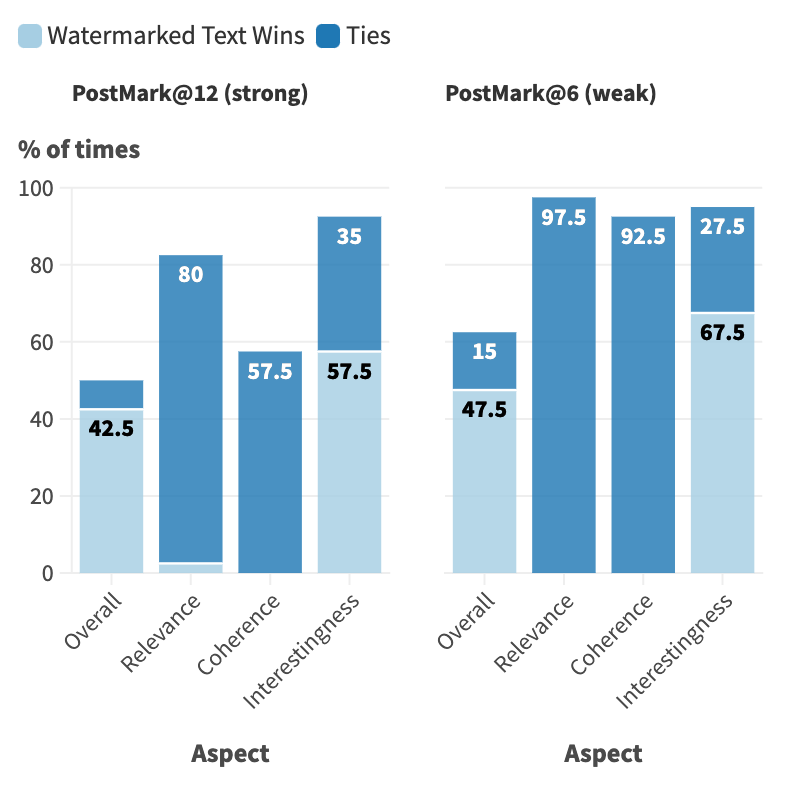}
\vspace{-0.15in}
\caption{Pairwise preference human evaluation results on \textsc{PostMark}@12 and \textsc{PostMark}@6. For both configurations, the watermarked text is at least as good as its unwatermarked counterpart the majority of the time in all aspects.}
\label{fig:pairwise-human}
\vspace{-0.15in}
\end{figure}

\paragraph{\textcolor{orange}{\emph{\textbf{> Q3:}}} \postmark watermarked texts are at least as good as their unwatermarked counterparts the majority of the time.} We first evaluate the impact of \postmark on quality through a pairwise comparison task, similar to the setup in~\autoref{sec:quality-auto-eval}. Each annotator reads 20 OpenGen prefixes and the corresponding pairs of anonymized watermarked and unwatermarked responses generated by \gpt. We then ask them to indicate their preferred response \textit{overall}, as well as their preferences in terms of \textit{relevance}, \textit{coherence}, and \textit{interestingness}, allowing for ties. Results in~\autoref{fig:pairwise-human} indicate that for \textsc{PostMark}@12 and \textsc{PostMark}@6, watermarked responses are at least as good as their unwatermarked counterparts the majority of the time (i.e., total percentage of wins and ties $\geq$ 50\%). As expected, reducing the insertion rate to 6\% improves quality, especially in the \textit{coherence} aspect.\footnote{While  soft win rates computed from human annotations are much lower than those from \gptturbo's judgments, both judges agree that a smaller $r$ improves quality.} To put things in perspective, a previous human evaluation study by~\citet{kirchenbauer2024on} found that annotators preferred KGW-watermarked text over unwatermarked text only 38.4\% of the time. 

\paragraph{\textcolor{orange}{\emph{\textbf{> Q4:}}} Annotators struggle to identify the words inserted by \postmark.} A primary concern with \postmark is whether the words inserted into the watermarked text will be conspicuous enough for humans to identify, making it easy for attackers to remove them. To measure this, we create an anonymized mixture of 20 unwatermarked\footnote{We include unwatermarked responses in this evaluation as a baseline. For fairness, we regenerated unwatermarked texts to roughly match the length of the watermarked texts.} and 20 watermarked responses generated for 20 prefixes in OpenGen with \gpt as the base LLM.\footnote{These 20 prefixes are different from the ones they see in the pairwise comparison evaluation.} We then ask annotators to highlight out-of-place words that they think might have been inserted post-hoc after the initial generation. Overall, annotators achieve a \textbf{F1} of merely \textbf{0.06} (0.46 precision, 0.03 recall). On average, they highlight 2.2 words in each unwatermarked response, and 3.45 words in each watermarked response. Thus, even when annotators are aware of the insertion of words, they cannot pinpoint the specific words.

\section{Related work} \label{sec:related-work}

\paragraph{Early research on watermarking:} Our work is relevant to early work on watermarking text documents, either using the text document image~\citep{brassil-1995-watermark, low-1998-centroid}, syntactic transformations~\citep{atallah-2001-syntax-watermark, meral-2009-syntax-watermark}, or semantic changes~\citep{atallah-2003-semantic-watermark, topraka-2006-synonym}. Later work also explores watermarking machine-generated text~\citep{venugopal-etal-2011-watermarking}.

\paragraph{Watermarking LLM-generated text:} Recent research has primarily focused on watermarking LLM-generated outputs. Most existing approaches operate in the \emph{whitebox} setting, assuming access to model logits and the ability to modify the decoding process~\citep{fang-etal-2017-generating, kaptchuk-2021-meteor, aaronson2022, pmlr-v202-kirchenbauer23a, zhao2023provable, liu2024an, liu2024a-sir} or inject detectable signals without altering the original token distribution~\citep{christ2023undetectable, kuditipudi2024robust}. Alternatively,~\citet{hou-etal-2023-semstamp, hou2024ksemstamp} watermark at the sentence level via rejection sampling. Prior \emph{blackbox} methods access only model outputs (like \postmark), but rely on simple lexical substitution~\citep{abdelnabi2021adversarial, qiang2023lex, yang2023blackbox, munyer2024deeptextmark}.

\paragraph{Evading watermark detection:} Our work also relates to prior work on text editing attacks designed to evade watermark detection.~\citet{he2024translation} propose a cross-lingual attack, while~\citet{kirchenbauer2024on} studies a copy-paste attack that embeds watermarked text into a larger human-written document.~\citet{krishna2023paraphrasing} train a controllable paraphraser that allows for control over lexical and syntactic diversity.~\citet{sadasivan2024aigenerated} design a recursive paraphrasing attack that repeatedly rewrites watermarked text. Similar to our work, several studies directly prompt an instruction-following LLM to paraphrase text~\citep{zhao2023provable, hou-etal-2023-semstamp, hou2024ksemstamp, liu2024a-sir, kirchenbauer2024on}. 

\paragraph{Quality-robustness trade-off:} Relevant to our discussion in~\autoref{sec:quality}, several recent papers highlight the impact of watermarking on quality. In line with our conclusions,~\citet{singh2023new} and~\citet{molenda2024waterjudge} both find that less robust watermarks tend to have less negative impact on text quality.

\section{Conclusion}\label{sec:conclusion}

We propose \postmark, a novel watermarking approach that only requires access to the underlying model's outputs, making it applicable by third-party entities to outputs from API providers. Through extensive experiments acorss eight baseline algorithms, five base LLMs, and three datasets, we show that \postmark is more robust to paraphrasing attacks than existing methods. We conduct a human evaluation to show that words inserted by \postmark are not easily identifiable by humans. We further run comprehensive quality evaluations covering \textit{coherence}, \textit{relevance}, \textit{interestingness}, and \textit{factuality}, and find that \postmark preserves text quality relatively well. Future work could look into further optimizing each of the three modules in \postmark, evaluating \postmark on attacks other than paraphrasing, or making logit-based methods less entropy-dependent.

\section*{Limitations} \label{sec:limitations}

In this section, we address the primary limitations of our work.

\paragraph{Other attacks:} Our work focuses on evaluating robustness of various watermarking methods against paraphrasing attacks. However, there are many other interesting and practical attacks that we do not consider, such as the copy-paste attack and the recursive paraphrasing attack discussed in~\autoref{sec:related-work}. We anticipate that \postmark will be less effective when the watermarked text is embedded in a larger human-written document or when it undergoes repeated paraphrasing, similar to other watermarking methods. We leave the exploration of these other types of attacks to future work.

\paragraph{Runtime and API costs:} The \postmark implementation used in all our main experiments relies on closed-source models from OpenAI (\openai and \gpto). As a result, the runtime and costs of running \postmark are heavily dependent on the API provider. Our cost estimate in~\S\ref{appendix-runtime} suggests that watermarking 100 tokens with the default \textsc{PostMark}@12 configuration costs around \$1.2 USD. However, the framework is highly flexible in terms of module selection. In fact, as demonstrated in \autoref{sec:exp-results}, an open-source implementation can perform nearly as well as the closed-source version. We leave the optimization of open-source implementations of \postmark to future work.

\section*{Ethical considerations} \label{sec:ethical}

Our human study was determined exempt by IRB review. All annotators have consented to the release of their annotations, and we ensured they were fairly compensated for their valuable contributions. Scientific artifacts are implemented for their intended usage. The risks associated with our framework are no greater than those already present in the large language models it utilizes~\citep{weidinger2021ethical}.

\section*{Acknowledgments}
We extend special gratitude to the Upwork annotators for their hard work. This project was partially supported by awards IIS-2202506 and IIS-2312949 from the National Science Foundation (NSF).

\bibliography{custom}

\begin{thebibliography}{55}
\providecommand{\natexlab}[1]{#1}

\bibitem[{Aaronson and Kirchner(2022)}]{aaronson2022}
Scott Aaronson and Hendrik Kirchner. 2022.
\newblock \href {https://www.scottaaronson.com/talks/watermark.ppt} {Watermarking gpt outputs}.

\bibitem[{Abdelnabi and Fritz(2021)}]{abdelnabi2021adversarial}
Sahar Abdelnabi and Mario Fritz. 2021.
\newblock \href {https://arxiv.org/abs/2009.03015} {Adversarial watermarking transformer: Towards tracing text provenance with data hiding}.
\newblock \emph{Preprint}, arXiv:2009.03015.

\bibitem[{AI@Meta(2024)}]{llama3modelcard}
AI@Meta. 2024.
\newblock \href {https://github.com/meta-llama/llama3/blob/main/MODEL_CARD.md} {Llama 3 model card}.

\bibitem[{Atallah et~al.(2001)Atallah, Raskin, Crogan, Hempelmann, Kerschbaum, Mohamed, and Naik}]{atallah-2001-syntax-watermark}
Mikhail~J. Atallah, Victor Raskin, Michael Crogan, Christian Hempelmann, Florian Kerschbaum, Dina Mohamed, and Sanket Naik. 2001.
\newblock Natural language watermarking: Design, analysis, and a proof-of-concept implementation.
\newblock In \emph{Information Hiding}, pages 185--200, Berlin, Heidelberg. Springer Berlin Heidelberg.

\bibitem[{Atallah et~al.(2003)Atallah, Raskin, Hempelmann, Karahan, Sion, Topkara, and Triezenberg}]{atallah-2003-semantic-watermark}
Mikhail~J. Atallah, Victor Raskin, Christian~F. Hempelmann, Mercan Karahan, Radu Sion, Umut Topkara, and Katrina~E. Triezenberg. 2003.
\newblock Natural language watermarking and tamperproofing.
\newblock In \emph{Information Hiding}, pages 196--212, Berlin, Heidelberg. Springer Berlin Heidelberg.

\bibitem[{Bai et~al.(2022)Bai, Jones, Ndousse, Askell, Chen, DasSarma, Drain, Fort, Ganguli, Henighan, Joseph, Kadavath, Kernion, Conerly, El-Showk, Elhage, Hatfield-Dodds, Hernandez, Hume, Johnston, Kravec, Lovitt, Nanda, Olsson, Amodei, Brown, Clark, McCandlish, Olah, Mann, and Kaplan}]{bai2022training}
Yuntao Bai, Andy Jones, Kamal Ndousse, Amanda Askell, Anna Chen, Nova DasSarma, Dawn Drain, Stanislav Fort, Deep Ganguli, Tom Henighan, Nicholas Joseph, Saurav Kadavath, Jackson Kernion, Tom Conerly, Sheer El-Showk, Nelson Elhage, Zac Hatfield-Dodds, Danny Hernandez, Tristan Hume, Scott Johnston, Shauna Kravec, Liane Lovitt, Neel Nanda, Catherine Olsson, Dario Amodei, Tom Brown, Jack Clark, Sam McCandlish, Chris Olah, Ben Mann, and Jared Kaplan. 2022.
\newblock \href {https://arxiv.org/abs/2204.05862} {Training a helpful and harmless assistant with reinforcement learning from human feedback}.
\newblock \emph{Preprint}, arXiv:2204.05862.

\bibitem[{Brassil et~al.(1995)Brassil, Low, Maxemchuk, and O'Gorman}]{brassil-1995-watermark}
J.T. Brassil, S.~Low, N.F. Maxemchuk, and L.~O'Gorman. 1995.
\newblock \href {https://doi.org/10.1109/49.464718} {Electronic marking and identification techniques to discourage document copying}.
\newblock \emph{IEEE Journal on Selected Areas in Communications}, 13(8):1495--1504.

\bibitem[{Christ et~al.(2023)Christ, Gunn, and Zamir}]{christ2023undetectable}
Miranda Christ, Sam Gunn, and Or~Zamir. 2023.
\newblock \href {https://arxiv.org/abs/2306.09194} {Undetectable watermarks for language models}.
\newblock \emph{Preprint}, arXiv:2306.09194.

\bibitem[{Computer(2023)}]{together2023redpajama}
Together Computer. 2023.
\newblock \href {https://github.com/togethercomputer/RedPajama-Data} {Redpajama: an open dataset for training large language models}.

\bibitem[{Fang et~al.(2017)Fang, Jaggi, and Argyraki}]{fang-etal-2017-generating}
Tina Fang, Martin Jaggi, and Katerina Argyraki. 2017.
\newblock \href {https://aclanthology.org/P17-3017} {Generating steganographic text with {LSTM}s}.
\newblock In \emph{Proceedings of {ACL} 2017, Student Research Workshop}, pages 100--106, Vancouver, Canada. Association for Computational Linguistics.

\bibitem[{He et~al.(2024)He, Zhou, Hao, Liu, Wang, Tu, Zhang, and Wang}]{he2024translation}
Zhiwei He, Binglin Zhou, Hongkun Hao, Aiwei Liu, Xing Wang, Zhaopeng Tu, Zhuosheng Zhang, and Rui Wang. 2024.
\newblock \href {https://arxiv.org/abs/2402.14007} {Can watermarks survive translation? on the cross-lingual consistency of text watermark for large language models}.
\newblock \emph{Preprint}, arXiv:2402.14007.

\bibitem[{Hou et~al.(2023)Hou, Zhang, He, Chuang, Wang, Shen, Van~Durme, Khashabi, and Tsvetkov}]{hou-etal-2023-semstamp}
Abe~Bohan* Hou, Jingyu* Zhang, Tianxing* He, Yung-Sung Chuang, Hongwei Wang, Lingfeng Shen, Benjamin Van~Durme, Daniel Khashabi, and Yulia Tsvetkov. 2023.
\newblock \href {https://arxiv.org/abs/2310.03991} {Semstamp: A semantic watermark with paraphrastic robustness for text generation}.
\newblock In \emph{Annual Conference of the North American Chapter of the Association for Computational Linguistics}.

\bibitem[{Hou et~al.(2024)Hou, Zhang, Wang, Khashabi, and He}]{hou2024ksemstamp}
Abe~Bohan Hou, Jingyu Zhang, Yichen Wang, Daniel Khashabi, and Tianxing He. 2024.
\newblock \href {https://arxiv.org/abs/2402.11399} {k-semstamp: A clustering-based semantic watermark for detection of machine-generated text}.
\newblock \emph{Preprint}, arXiv:2402.11399.

\bibitem[{Hu et~al.(2024)Hu, Chen, Wu, Wu, Zhang, and Huang}]{hu2024unbiased}
Zhengmian Hu, Lichang Chen, Xidong Wu, Yihan Wu, Hongyang Zhang, and Heng Huang. 2024.
\newblock \href {https://openreview.net/forum?id=uWVC5FVidc} {Unbiased watermark for large language models}.
\newblock In \emph{The Twelfth International Conference on Learning Representations}.

\bibitem[{Jiang et~al.(2023)Jiang, Sablayrolles, Mensch, Bamford, Chaplot, de~las Casas, Bressand, Lengyel, Lample, Saulnier, Lavaud, Lachaux, Stock, Scao, Lavril, Wang, Lacroix, and Sayed}]{jiang2023mistral}
Albert~Q. Jiang, Alexandre Sablayrolles, Arthur Mensch, Chris Bamford, Devendra~Singh Chaplot, Diego de~las Casas, Florian Bressand, Gianna Lengyel, Guillaume Lample, Lucile Saulnier, Lélio~Renard Lavaud, Marie-Anne Lachaux, Pierre Stock, Teven~Le Scao, Thibaut Lavril, Thomas Wang, Timothée Lacroix, and William~El Sayed. 2023.
\newblock \href {https://arxiv.org/abs/2310.06825} {Mistral 7b}.
\newblock \emph{Preprint}, arXiv:2310.06825.

\bibitem[{Kaptchuk et~al.(2021)Kaptchuk, Jois, Green, and Rubin}]{kaptchuk-2021-meteor}
Gabriel Kaptchuk, Tushar~M. Jois, Matthew Green, and Aviel~D. Rubin. 2021.
\newblock \href {https://doi.org/10.1145/3460120.3484550} {Meteor: Cryptographically secure steganography for realistic distributions}.
\newblock In \emph{Proceedings of the 2021 ACM SIGSAC Conference on Computer and Communications Security}, CCS '21, page 1529–1548, New York, NY, USA. Association for Computing Machinery.

\bibitem[{Kirchenbauer et~al.(2023)Kirchenbauer, Geiping, Wen, Katz, Miers, and Goldstein}]{pmlr-v202-kirchenbauer23a}
John Kirchenbauer, Jonas Geiping, Yuxin Wen, Jonathan Katz, Ian Miers, and Tom Goldstein. 2023.
\newblock \href {https://proceedings.mlr.press/v202/kirchenbauer23a.html} {A watermark for large language models}.
\newblock In \emph{Proceedings of the 40th International Conference on Machine Learning}, volume 202 of \emph{Proceedings of Machine Learning Research}, pages 17061--17084. PMLR.

\bibitem[{Kirchenbauer et~al.(2024)Kirchenbauer, Geiping, Wen, Shu, Saifullah, Kong, Fernando, Saha, Goldblum, and Goldstein}]{kirchenbauer2024on}
John Kirchenbauer, Jonas Geiping, Yuxin Wen, Manli Shu, Khalid Saifullah, Kezhi Kong, Kasun Fernando, Aniruddha Saha, Micah Goldblum, and Tom Goldstein. 2024.
\newblock \href {https://openreview.net/forum?id=DEJIDCmWOz} {On the reliability of watermarks for large language models}.
\newblock In \emph{The Twelfth International Conference on Learning Representations}.

\bibitem[{Krishna et~al.(2023)Krishna, Song, Karpinska, Wieting, and Iyyer}]{krishna2023paraphrasing}
Kalpesh Krishna, Yixiao Song, Marzena Karpinska, John Wieting, and Mohit Iyyer. 2023.
\newblock \href {https://openreview.net/forum?id=WbFhFvjjKj} {Paraphrasing evades detectors of {AI}-generated text, but retrieval is an effective defense}.
\newblock In \emph{Thirty-seventh Conference on Neural Information Processing Systems}.

\bibitem[{Kuditipudi et~al.(2024)Kuditipudi, Thickstun, Hashimoto, and Liang}]{kuditipudi2024robust}
Rohith Kuditipudi, John Thickstun, Tatsunori Hashimoto, and Percy Liang. 2024.
\newblock \href {https://arxiv.org/abs/2307.15593} {Robust distortion-free watermarks for language models}.
\newblock \emph{Preprint}, arXiv:2307.15593.

\bibitem[{Lambert et~al.(2024)Lambert, Pyatkin, Morrison, Miranda, Lin, Chandu, Dziri, Kumar, Zick, Choi, Smith, and Hajishirzi}]{lambert2024rewardbench}
Nathan Lambert, Valentina Pyatkin, Jacob Morrison, LJ~Miranda, Bill~Yuchen Lin, Khyathi Chandu, Nouha Dziri, Sachin Kumar, Tom Zick, Yejin Choi, Noah~A. Smith, and Hannaneh Hajishirzi. 2024.
\newblock \href {https://arxiv.org/abs/2403.13787} {Rewardbench: Evaluating reward models for language modeling}.
\newblock \emph{Preprint}, arXiv:2403.13787.

\bibitem[{Liu et~al.(2024{\natexlab{a}})Liu, Pan, Hu, Li, Wen, King, and Yu}]{liu2024an}
Aiwei Liu, Leyi Pan, Xuming Hu, Shuang Li, Lijie Wen, Irwin King, and Philip~S. Yu. 2024{\natexlab{a}}.
\newblock \href {https://openreview.net/forum?id=gMLQwKDY3N} {An unforgeable publicly verifiable watermark for large language models}.
\newblock In \emph{The Twelfth International Conference on Learning Representations}.

\bibitem[{Liu et~al.(2024{\natexlab{b}})Liu, Pan, Hu, Meng, and Wen}]{liu2024a-sir}
Aiwei Liu, Leyi Pan, Xuming Hu, Shiao Meng, and Lijie Wen. 2024{\natexlab{b}}.
\newblock \href {https://openreview.net/forum?id=6p8lpe4MNf} {A semantic invariant robust watermark for large language models}.
\newblock In \emph{The Twelfth International Conference on Learning Representations}.

\bibitem[{Low et~al.(1998)Low, Maxemchuk, and Lapone}]{low-1998-centroid}
S.H. Low, N.F. Maxemchuk, and A.M. Lapone. 1998.
\newblock \href {https://doi.org/10.1109/26.662643} {Document identification for copyright protection using centroid detection}.
\newblock \emph{IEEE Transactions on Communications}, 46(3):372--383.

\bibitem[{Meral et~al.(2009)Meral, Sankur, {Sumru Özsoy}, Güngör, and Sevinç}]{meral-2009-syntax-watermark}
Hasan~Mesut Meral, Bülent Sankur, A.~{Sumru Özsoy}, Tunga Güngör, and Emre Sevinç. 2009.
\newblock \href {https://doi.org/10.1016/j.csl.2008.04.001} {Natural language watermarking via morphosyntactic alterations}.
\newblock \emph{Computer Speech and Language}, 23(1):107--125.

\bibitem[{Merity et~al.(2017)Merity, Xiong, Bradbury, and Socher}]{merity2017pointer}
Stephen Merity, Caiming Xiong, James Bradbury, and Richard Socher. 2017.
\newblock \href {https://openreview.net/forum?id=Byj72udxe} {Pointer sentinel mixture models}.
\newblock In \emph{International Conference on Learning Representations}.

\bibitem[{Min et~al.(2023)Min, Krishna, Lyu, Lewis, Yih, Koh, Iyyer, Zettlemoyer, and Hajishirzi}]{min-etal-2023-factscore}
Sewon Min, Kalpesh Krishna, Xinxi Lyu, Mike Lewis, Wen-tau Yih, Pang Koh, Mohit Iyyer, Luke Zettlemoyer, and Hannaneh Hajishirzi. 2023.
\newblock \href {https://doi.org/10.18653/v1/2023.emnlp-main.741} {{FA}ct{S}core: Fine-grained atomic evaluation of factual precision in long form text generation}.
\newblock In \emph{Proceedings of the 2023 Conference on Empirical Methods in Natural Language Processing}, pages 12076--12100, Singapore. Association for Computational Linguistics.

\bibitem[{Mitchell et~al.(2023)Mitchell, Lee, Khazatsky, Manning, and Finn}]{mitchell2023detectgpt}
Eric Mitchell, Yoonho Lee, Alexander Khazatsky, Christopher~D. Manning, and Chelsea Finn. 2023.
\newblock \href {https://arxiv.org/abs/2301.11305} {Detectgpt: Zero-shot machine-generated text detection using probability curvature}.
\newblock \emph{Preprint}, arXiv:2301.11305.

\bibitem[{Molenda et~al.(2024)Molenda, Liusie, and Gales}]{molenda2024waterjudge}
Piotr Molenda, Adian Liusie, and Mark J.~F. Gales. 2024.
\newblock \href {https://arxiv.org/abs/2403.19548} {Waterjudge: Quality-detection trade-off when watermarking large language models}.
\newblock \emph{Preprint}, arXiv:2403.19548.

\bibitem[{Muennighoff et~al.(2023)Muennighoff, Tazi, Magne, and Reimers}]{muennighoff-etal-2023-mteb}
Niklas Muennighoff, Nouamane Tazi, Loic Magne, and Nils Reimers. 2023.
\newblock \href {https://doi.org/10.18653/v1/2023.eacl-main.148} {{MTEB}: Massive text embedding benchmark}.
\newblock In \emph{Proceedings of the 17th Conference of the European Chapter of the Association for Computational Linguistics}, pages 2014--2037, Dubrovnik, Croatia. Association for Computational Linguistics.

\bibitem[{Munyer et~al.(2024)Munyer, Tanvir, Das, and Zhong}]{munyer2024deeptextmark}
Travis Munyer, Abdullah Tanvir, Arjon Das, and Xin Zhong. 2024.
\newblock \href {https://arxiv.org/abs/2305.05773} {Deeptextmark: A deep learning-driven text watermarking approach for identifying large language model generated text}.
\newblock \emph{Preprint}, arXiv:2305.05773.

\bibitem[{Navigli et~al.(2023)Navigli, Conia, and Ross}]{navigli2023biases}
Roberto Navigli, Simone Conia, and Bj\"{o}rn Ross. 2023.
\newblock \href {https://doi.org/10.1145/3597307} {Biases in large language models: Origins, inventory, and discussion}.
\newblock \emph{J. Data and Information Quality}, 15(2).

\bibitem[{Nussbaum et~al.(2024)Nussbaum, Morris, Duderstadt, and Mulyar}]{nussbaum2024nomic}
Zach Nussbaum, John~X. Morris, Brandon Duderstadt, and Andriy Mulyar. 2024.
\newblock \href {https://arxiv.org/abs/2402.01613} {Nomic embed: Training a reproducible long context text embedder}.
\newblock \emph{Preprint}, arXiv:2402.01613.

\bibitem[{{OpenAI}()}]{openai-gpt4o}
{OpenAI}.
\newblock \href {https://openai.com/index/hello-gpt-4o/} {{Model release blog: GPT-4o}}.
\newblock Technical report, OpenAI.

\bibitem[{OpenAI(2024{\natexlab{a}})}]{openai2024gpt4}
OpenAI. 2024{\natexlab{a}}.
\newblock \href {https://arxiv.org/abs/2303.08774} {Gpt-4 technical report}.
\newblock \emph{Preprint}, arXiv:2303.08774.

\bibitem[{OpenAI(2024{\natexlab{b}})}]{2024openaiemb}
OpenAI. 2024{\natexlab{b}}.
\newblock \href {https://openai.com/index/new-embedding-models-and-api-updates/} {New embedding models and api updates}.

\bibitem[{Pan et~al.(2024)Pan, Liu, He, Gao, Zhao, Lu, Zhou, Liu, Hu, Wen, and King}]{pan2024markllm}
Leyi Pan, Aiwei Liu, Zhiwei He, Zitian Gao, Xuandong Zhao, Yijian Lu, Binglin Zhou, Shuliang Liu, Xuming Hu, Lijie Wen, and Irwin King. 2024.
\newblock \href {https://arxiv.org/abs/2405.10051} {Markllm: An open-source toolkit for llm watermarking}.
\newblock \emph{Preprint}, arXiv:2405.10051.

\bibitem[{Panickssery et~al.(2024)Panickssery, Bowman, and Feng}]{panickssery2024llm}
Arjun Panickssery, Samuel~R. Bowman, and Shi Feng. 2024.
\newblock \href {https://arxiv.org/abs/2404.13076} {Llm evaluators recognize and favor their own generations}.
\newblock \emph{Preprint}, arXiv:2404.13076.

\bibitem[{Qiang et~al.(2023)Qiang, Zhu, Li, Zhu, Yuan, and Wu}]{qiang2023lex}
Jipeng Qiang, Shiyu Zhu, Yun Li, Yi~Zhu, Yunhao Yuan, and Xindong Wu. 2023.
\newblock \href {https://doi.org/10.1016/j.artint.2023.103859} {Natural language watermarking via paraphraser-based lexical substitution}.
\newblock \emph{Artif. Intell.}, 317(C).

\bibitem[{Raffel et~al.(2020)Raffel, Shazeer, Roberts, Lee, Narang, Matena, Zhou, Li, and Liu}]{raffel2020c4}
Colin Raffel, Noam Shazeer, Adam Roberts, Katherine Lee, Sharan Narang, Michael Matena, Yanqi Zhou, Wei Li, and Peter~J. Liu. 2020.
\newblock Exploring the limits of transfer learning with a unified text-to-text transformer.
\newblock \emph{J. Mach. Learn. Res.}, 21(1).

\bibitem[{Sadasivan et~al.(2024)Sadasivan, Kumar, Balasubramanian, Wang, and Feizi}]{sadasivan2024aigenerated}
Vinu~Sankar Sadasivan, Aounon Kumar, Sriram Balasubramanian, Wenxiao Wang, and Soheil Feizi. 2024.
\newblock \href {https://arxiv.org/abs/2303.11156} {Can ai-generated text be reliably detected?}
\newblock \emph{Preprint}, arXiv:2303.11156.

\bibitem[{Shumailov et~al.(2023)Shumailov, Shumaylov, Zhao, Gal, Papernot, and Anderson}]{shumailov2023curse}
Ilia Shumailov, Zakhar Shumaylov, Yiren Zhao, Yarin Gal, Nicolas Papernot, and Ross Anderson. 2023.
\newblock \href {https://arxiv.org/abs/2305.17493} {The curse of recursion: Training on generated data makes models forget}.
\newblock \emph{Preprint}, arXiv:2305.17493.

\bibitem[{Singh and Zou(2023)}]{singh2023new}
Karanpartap Singh and James Zou. 2023.
\newblock \href {https://arxiv.org/abs/2312.02382} {New evaluation metrics capture quality degradation due to llm watermarking}.
\newblock \emph{Preprint}, arXiv:2312.02382.

\bibitem[{Tian(2023)}]{2023gptzero}
Edward Tian. 2023.
\newblock \href {https://gptzero.me/} {Gptzero: An ai text detector}.

\bibitem[{Topkara et~al.(2006)Topkara, Topkara, and Atallah}]{topraka-2006-synonym}
Umut Topkara, Mercan Topkara, and Mikhail~J. Atallah. 2006.
\newblock \href {https://doi.org/10.1145/1161366.1161397} {The hiding virtues of ambiguity: quantifiably resilient watermarking of natural language text through synonym substitutions}.
\newblock In \emph{Proceedings of the 8th Workshop on Multimedia and Security}, page 164–174, New York, NY, USA. Association for Computing Machinery.

\bibitem[{Venugopal et~al.(2011)Venugopal, Uszkoreit, Talbot, Och, and Ganitkevitch}]{venugopal-etal-2011-watermarking}
Ashish Venugopal, Jakob Uszkoreit, David Talbot, Franz Och, and Juri Ganitkevitch. 2011.
\newblock \href {https://aclanthology.org/D11-1126} {Watermarking the outputs of structured prediction with an application in statistical machine translation.}
\newblock In \emph{Proceedings of the 2011 Conference on Empirical Methods in Natural Language Processing}, pages 1363--1372, Edinburgh, Scotland, UK. Association for Computational Linguistics.

\bibitem[{Wang et~al.(2023)Wang, Li, Chen, Cai, Zhu, Lin, Cao, Liu, Liu, and Sui}]{wang2023large}
Peiyi Wang, Lei Li, Liang Chen, Zefan Cai, Dawei Zhu, Binghuai Lin, Yunbo Cao, Qi~Liu, Tianyu Liu, and Zhifang Sui. 2023.
\newblock \href {https://arxiv.org/abs/2305.17926} {Large language models are not fair evaluators}.
\newblock \emph{Preprint}, arXiv:2305.17926.

\bibitem[{Wang et~al.(2022)Wang, Deng, Sun, and Meng}]{Wang2022PerplexityFP}
Yequan Wang, Jiawen Deng, Aixin Sun, and Xuying Meng. 2022.
\newblock \href {https://api.semanticscholar.org/CorpusID:265095122} {Perplexity from plm is unreliable for evaluating text quality}.

\bibitem[{Weidinger et~al.(2021)Weidinger, Mellor, Rauh, Griffin, Uesato, Huang, Cheng, Glaese, Balle, Kasirzadeh, Kenton, Brown, Hawkins, Stepleton, Biles, Birhane, Haas, Rimell, Hendricks, Isaac, Legassick, Irving, and Gabriel}]{weidinger2021ethical}
Laura Weidinger, John Mellor, Maribeth Rauh, Conor Griffin, Jonathan Uesato, Po-Sen Huang, Myra Cheng, Mia Glaese, Borja Balle, Atoosa Kasirzadeh, Zac Kenton, Sasha Brown, Will Hawkins, Tom Stepleton, Courtney Biles, Abeba Birhane, Julia Haas, Laura Rimell, Lisa~Anne Hendricks, William Isaac, Sean Legassick, Geoffrey Irving, and Iason Gabriel. 2021.
\newblock \href {https://arxiv.org/abs/2112.04359} {Ethical and social risks of harm from language models}.
\newblock \emph{Preprint}, arXiv:2112.04359.

\bibitem[{Wieting et~al.(2015)Wieting, Bansal, Gimpel, and Livescu}]{wieting-etal-2015-paraphrase}
John Wieting, Mohit Bansal, Kevin Gimpel, and Karen Livescu. 2015.
\newblock \href {https://doi.org/10.1162/tacl_a_00143} {From paraphrase database to compositional paraphrase model and back}.
\newblock \emph{Transactions of the Association for Computational Linguistics}, 3:345--358.

\bibitem[{Xu et~al.(2024)Xu, Jain, and Kankanhalli}]{xu2024hallucination}
Ziwei Xu, Sanjay Jain, and Mohan Kankanhalli. 2024.
\newblock \href {https://arxiv.org/abs/2401.11817} {Hallucination is inevitable: An innate limitation of large language models}.
\newblock \emph{Preprint}, arXiv:2401.11817.

\bibitem[{Yang et~al.(2023)Yang, Chen, Zhang, Liu, Qi, Zhang, Fang, and Yu}]{yang2023blackbox}
Xi~Yang, Kejiang Chen, Weiming Zhang, Chang Liu, Yuang Qi, Jie Zhang, Han Fang, and Nenghai Yu. 2023.
\newblock \href {https://arxiv.org/abs/2305.08883} {Watermarking text generated by black-box language models}.
\newblock \emph{Preprint}, arXiv:2305.08883.

\bibitem[{Zhang et~al.(2022)Zhang, Roller, Goyal, Artetxe, Chen, Chen, Dewan, Diab, Li, Lin, Mihaylov, Ott, Shleifer, Shuster, Simig, Koura, Sridhar, Wang, and Zettlemoyer}]{zhang2022opt}
Susan Zhang, Stephen Roller, Naman Goyal, Mikel Artetxe, Moya Chen, Shuohui Chen, Christopher Dewan, Mona Diab, Xian Li, Xi~Victoria Lin, Todor Mihaylov, Myle Ott, Sam Shleifer, Kurt Shuster, Daniel Simig, Punit~Singh Koura, Anjali Sridhar, Tianlu Wang, and Luke Zettlemoyer. 2022.
\newblock \href {https://arxiv.org/abs/2205.01068} {Opt: Open pre-trained transformer language models}.
\newblock \emph{Preprint}, arXiv:2205.01068.

\bibitem[{Zhao et~al.(2023)Zhao, Ananth, Li, and Wang}]{zhao2023provable}
Xuandong Zhao, Prabhanjan Ananth, Lei Li, and Yu-Xiang Wang. 2023.
\newblock \href {https://arxiv.org/abs/2306.17439} {Provable robust watermarking for ai-generated text}.
\newblock \emph{Preprint}, arXiv:2306.17439.

\bibitem[{Zheng et~al.(2023)Zheng, Chiang, Sheng, Zhuang, Wu, Zhuang, Lin, Li, Li, Xing, Zhang, Gonzalez, and Stoica}]{zheng2023judging}
Lianmin Zheng, Wei-Lin Chiang, Ying Sheng, Siyuan Zhuang, Zhanghao Wu, Yonghao Zhuang, Zi~Lin, Zhuohan Li, Dacheng Li, Eric~P. Xing, Hao Zhang, Joseph~E. Gonzalez, and Ion Stoica. 2023.
\newblock \href {https://arxiv.org/abs/2306.05685} {Judging llm-as-a-judge with mt-bench and chatbot arena}.
\newblock \emph{Preprint}, arXiv:2306.05685.

\end{thebibliography}

\newpage

\appendix
\section{More details on the vocabulary $\mathbb{V}$ of the \sectable} \label{appendix-sectable}

In this section, we provide more details on the creation of \sectable, and address how often a word in the \sectable can be selected as a watermark word.

\paragraph{Filtering the \sectable vocabulary $\mathbb{V}$:}  Specifically, we restrict $\mathbb{V}$ to only include lowercase nouns, verbs, adjectives, and adverbs that occur at least 1,000 times in the WikiText-103 training split. This results in a final vocabulary of 3,266 words.

\begin{figure}[htbp]
    \centering
    \includegraphics[scale=0.25]{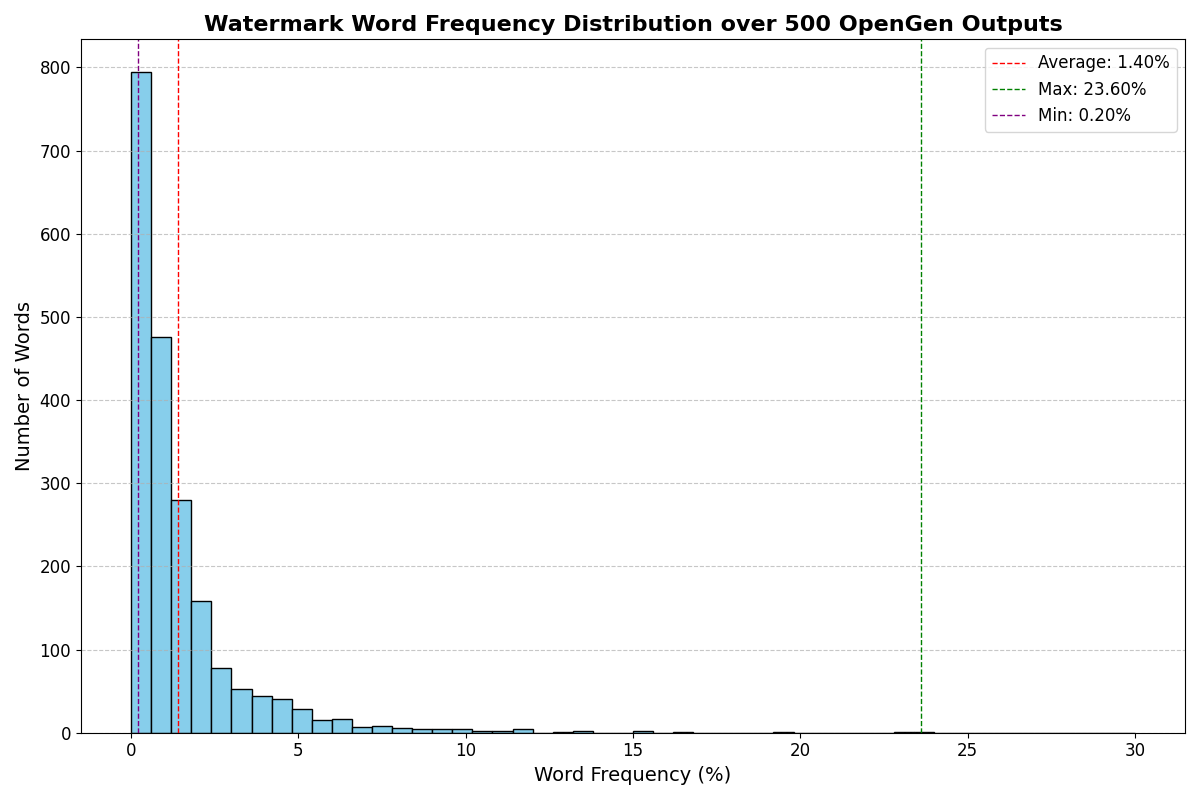}
    \caption{Watermark word frequency distribution over 500 OpenGen outputs. The majority of the words are chosen as watermark words less than 5\% of the time. There are only two major hub words that are selected more than 20\% of the time.}
    \label{fig:word-freq}
\end{figure}

\paragraph{Frequency of words chosen as watermark words:} In~\autoref{fig:word-freq}, we plot the frequency distribution of all watermark words obtained for 500 OpenGen outputs (generated with \gpt as the base LLM). We find that the majority of the words are only selected as watermark words for less than 5\% of all outputs, while two major hub words are selected in more than 20\% of the outputs. Overall, the hubness problem is not too severe, but it could be mitigated by a more careful selection of the embeddings used in the \sectable.

\section{Prompt for the \ins} \label{appendix-insertion-prompt}

\begin{spverbatim}
Given below are a piece of text and a word list. Rewrite the text to incorporate all words from the provided word list. The rewritten text must be coherent and factual. Distribute the words from the list evenly throughout the text, rather than clustering them in a single section. When rewriting the text, try your best to minimize text length increase. Only return the rewritten text in your response, do not say anything else.

Text: {}

Word list: {}

Rewritten text:
\end{spverbatim}

\section{More details on cosine similarity word matching during detection} \label{appendix-cos-sim-detection}

We use the \paragram\ word embedding model developed by~\citet{wieting-etal-2015-paraphrase} to perform cosine similarity word matching during detection. We find this model to be superior at distinguishing semantically related words from irrelevant words, see details in~\autoref{tab:paragram-sim}.

\begin{table}[]
\scriptsize
\centering
\begin{tabular}{lcc}
\toprule
         & SIM(positive) & SIM(negative) \\
\midrule
\paragram\ & 64.8         & 2.4          \\
\textsc{GloVe}    & 60.7         & 16.4         \\
\nomic\    & 59.9         & 33.2         \\
\openai\   & 64.2         & 29.8        \\
\bottomrule
\end{tabular}
\caption{Cosine similarity between embeddings of positive pairs (word + its synonym) and between negative pairs (word + irrelevant word) computed with different embedding models, averaged over 174 tuples of (word, synonym, irrelevant word).}
\label{tab:paragram-sim}
\end{table}

\section{More details on baselines} \label{appendix-baseline}

In this section, we provide more details on how we run our baselines.

\subsection{Expanded descriptions of baselines} \label{appendix-baseline-desc}

(1) \textbf{KGW}~\citep{pmlr-v202-kirchenbauer23a}: Partitions the vocabulary into ``green'' and ``red'' lists based on the previous token, then boosts the probability of green tokens during generation. Detection is done by comparing the number of green tokens present to the expected count under the null hypothesis of no watermarking. (2) \textbf{Unigram}~\citep{zhao2023provable}: A variant of KGW that uses a fixed green-red partition for all generations instead of re-partitioning the vocabulary at each token, making it more robust to editing attacks. (3) \textbf{EXP}~\citep{aaronson2022}: Uses exponential sampling to embed a watermark by biasing token selection with a pseudo-random sequence during text generation. Detection measures the correlation between the generated text and the sequence to identify the watermark. (4) \textbf{EXP-Edit}~\citep{kuditipudi2024robust}: A variant of the EXP watermark that incorporates edit distance to measure the correlation. (5) \textbf{SemStamp}~\citep{hou-etal-2023-semstamp}: A sentence-level algorithm that partitions the semantic space using locality-sensitive hashing with arbitrary hyperplanes, assigning binary signatures to regions and accepting sentences that fall within ``valid'' regions, which enhances robustness against paraphrase attacks. (6) \textbf{k-SemStamp}~\citep{hou2024ksemstamp}: Improves upon SemStamp by using k-means clustering to partition the semantic space. (7) \textbf{SIR}~\citep{liu2024a-sir}: Generates watermark logits from the semantic embeddings of preceding tokens using an embedding language model and a trained watermark model. These logits are added to the language model's logits. Detection works by averaging these watermark logits for each token and identifying a watermark if the average is significantly greater than zero. (8) \textbf{Blackbox}~\citep{yang2023blackbox}: While all other baseline methods require access to model logits, this method focuses on the blackbox setting where only the model output is observable, similar to our assumption. It encodes words as binary bits, replaces bit-0 words with synonyms representing bit-1, and detects watermarks through a statistical test identifying the altered distribution of binary bits.

\subsection{Hyperparameters for baselines} \label{appendix-baseline-hparam}

All baselines are run with nucleus sampling with $p = 0.9$ unless otherwise specified.

\paragraph{KGW:} We run KGW in the LeftHash configuration with $\gamma=0.5$ and $\delta=4.0$, using the original authors' implementation. These hyper-parameters control the size of the green token list and the strength of the watermark, respectively. While $\delta$ is typically set to $2.0$ in prior literature, we chose $\delta=4.0$ based on findings by \citet{kirchenbauer2024on}. They found that $\delta=4.0$ made the watermark more robust to paraphrasing attacks in their experiments with Vicuna, a supervised instruction-finetuned model. Given that our experiments also focus on lower-entropy models aligned through RLHF or instruction tuning, we adopt the same value for $\delta$.

\paragraph{Unigram:} To align with the setup of KGW, we set $\gamma=0.5$ and $\delta=4.0$ for Unigram as well. While the authors open-source their code, we ran into unexpected performance issues, where Unigram could not even achieve a TPR at 1\% FPR higher than 70\% even before any attacks on OpenGen with \llama as the base model. Thus, we switched to the implementation in MarkLLM~\citep{pan2024markllm}, an open-source watermarking toolkit. With this implementation, Unigram's TPR before attacks became close to 100\% and the TPR after attacks stayed above 90\%, in line with results reported in the Unigram paper~\citep{zhao2023provable}.

\paragraph{EXP:} We run EXP with prefix length set to $1$ using the MarkLLM implementation.

\paragraph{EXP-Edit:} Using the authors' implementation, we run EXP-Edit with $\gamma = 0.5$, watermark key length = 256, block size = sequence length = 300, and number of resamples = 100. This method is run with multinomial sampling (the default setting in the authors' code), because we find that adding a nucleus sampling logits wrapper on top significantly hurts its performance. For \llamainst and \mistral, we find that this method cannot reach a TPR at 1\% FPR above 70\% even before attacks. We tried several values for $\gamma$, the hyperparameter that controls the statistical power of the watermark, but it did not improve the results. Increasing the number of resamples to 500 also had little effect.

\paragraph{Blackbox:} We run Blackbox with $\tau = 0.8$ and $\lambda = 0.83$ using fast detection with the authors' implementation. Empirically, we find that fast detection offers a significant speed advantage with negligible impact on performance when compared to precise detection. On \textbf{200} OpenGen outputs with \gpt as the base LLM, using precise detection yields TPR of \textcolor{purple}{100} before paraphrasing and \textcolor{teal}{3.5} after paraphrasing, whereas fast detection yields \textcolor{purple}{99} and \textcolor{teal}{0.5}.

\section{More details on base models} \label{appendix-base-models}

In this section, we provide more details on how we run the base generator models.

\paragraph{Model checkpoints:}

We detail the checkpoint we use for each base model in~\autoref{tab:base-model-ckpt}.

\begin{table}[]
\small
\centering
\begin{tabular}{ll}
\toprule
Model & Checkpoint \\
\midrule
\llama     & {Meta-Llama-3-8B} \\
\llamainst & {Meta-Llama-3-8B-Instruct} \\
\mistral   & {Mistral-7B-Instruct-v0.2}  \\
\gpt       & OpenAI API (\texttt{gpt-4-0613}) \\
\bottomrule
\end{tabular}
\caption{Base model checkpoints.}
\label{tab:base-model-ckpt}
\end{table}


\paragraph{Generation length:} For all aligned models (\llamainst, \mistral, and \gpt), we generate free-form text until the model outputs an EOS (end-of-sequence) token to simulate the downstream setting. For \llama, we set the maximum token limit to 300, as generating freely until reaching EOS often leads to meaningless repetitions, sometimes even exceeding 8,000 tokens. We do not run \opt ourselves.

\section{Paraphrasing attack setup} \label{appendix-para}

In this section, we provide more details on the paraphrasing attack we use for all experiments.

\paragraph{Prompt for sentence-level paraphrasing:} We build on the prompt used by~\citet{hou-etal-2023-semstamp, hou2024ksemstamp} and include more clarification on what to return:

\begin{spverbatim}
Given some previous context and a sentence following that context, paraphrase the current sentence. Only return the paraphrased sentence in your response.

Previous context: {}
Current sentence to paraphrase: {}
Your paraphrase of the current sentence: 
\end{spverbatim}

\paragraph{Why sentence-level paraphrasing?} We choose a sentence-level paraphrasing setup for two reasons. First,~\citet{hou-etal-2023-semstamp, hou2024ksemstamp} use a sentence-level paraphrasing setup to evaluate the robustness of their method. Since we are unable to run their method directly, adopting the same paraphrasing setup allows for a fair comparison with their results. Second, as observed by~\citet{kirchenbauer2024on}, naively prompting \gpttturbo to rewrite the entire input text often results in significant loss of important content. While the authors developed a sophisticated prompt to mitigate this issue, we empirically find that paraphrasing at a sentence level achieves a similar effect.

\section{\postmark runtime and API cost estimates} \label{appendix-runtime}

\paragraph{Runtime:} We compare the runtime of several \postmark\ configurations with other baselines in~\autoref{tab:runtime}. Recall that in our experiments, we find insertion success rate to be higher if we divide the watermark word list into sublists of 10 words, then ask the \ins\ to insert one sublist at a time. This iterative insertion process can have some negative impact on runtime, but it may become unnecessary in the future when the \ins\ has better instruction-following capabilities.

\paragraph{API costs:} Under the default \textsc{PostMark}@12 configuration with \gpto\ as the \ins\ and \openai\ as the \emb\, watermarking 500 outputs with around 300 tokens costs around \$18.5 USD, which means that watermarking 100 tokens costs about \$1.2 on average.

\begin{table}[]
\small
\centering
\begin{tabular}{lc}
\toprule
Method & Avg Time / Output \\
\midrule
\textsc{PostMark}@6 & 29.2 \\
\textsc{PostMark}@12 & 36.6 \\
\textsc{PostMark}@12 (no iter.) & 25.3 \\
KGW      & 17.5 \\
Unigram  & 18.5 \\
EXP      & 18.4 \\
EXP-Edit & 17.3 \\
Blackbox & 21.6 \\
\bottomrule
\end{tabular}
\caption{Average time (in seconds) it takes to generate one watermarked instance with \llamainst as the base LLM. Runtime is averaged over 10 outputs, with an average token count of 280. For \postmark and Blackbox, the runtime includes the time it takes for \llamainst to generate the initial unwatermarked output. \textsc{PostMark}@12 (no iter.) refers to the setup where instead of breaking up the watermark word list into sublists and iteratively asking the \ins to insert one sublist at a time, we directly ask the \ins to insert all words in the list.}
\label{tab:runtime}
\end{table}

\section{\postmark length comparison} \label{appendix-length}

We present a comparison between output length (before and after watermarking) for various watermarking methods in~\autoref{tab:length}.

\begin{table*}[ht]
\renewcommand{\arraystretch}{1.2}
\centering
\scriptsize
\begin{tabular}{lcccccc}
\toprule
Metric $\rightarrow$ & \multicolumn{6}{c}{ Number of Tokens (Before / After Watermarking) } \\
\cline{2-7}
Methods $\rightarrow$ & KGW & Unigram & EXP & EXP-Edit & Blackbox & \textsc{PostMark}@12 \\
\midrule 
\llama & 239.6 / 226.6 & 237.6 / 250.7 & 232.5 / 269.8 & 213 / 225.7 & 239.6 / 244.8 & 239.6 / 381.2 \\
\midrule 
\llamainst  & 251.2 / 234.6 & 259.5 / 261.6 & 259 / 282.6 & 251.3 / 255 & 251.2 / 256.4 & 251.2 / 431 \\
\midrule 
\mistral & 315.3 / 588.2 & 318 / 321 & 317.4 / 247.8 & 248.7 / 249.5 & 315.3 / 320.6 & 315.3 / 552.2 \\
\midrule 
\gpt & - & - & - & - & 301.2 / 305.7 & 301.2 / 507.1 \\
\bottomrule 
\end{tabular}
\caption{Length comparison between different watermarking methods before and after watermarking, averaged over 500 OpenGen outputs.}
\label{tab:length}
\end{table*}

\section{Unigram repetitions} \label{appendix-unigram-rep}

We present several examples of Unigram's repetitive watermarked outputs in~\autoref{tab:unigram-outputs}, generated with \llama as the base LLM.

\begin{table*}[]
\small
\centering
\begin{tabular}{p{0.2\textwidth} p{0.35\textwidth} p{0.35\textwidth}}
\toprule
Prefix & Without Watermarking & After Unigram Watermarking \\
\midrule
Unlike mountains of similar altitude elsewhere, Elbert lacks both a permanent snowpack and a prominent north-facing cirque, which can be attributed to its position among other mountains of similar height, causing it to receive relatively small quantities of precipitation. Mount Elbert was named by miners in honor of Samuel Hitt Elbert, the governor of the then-Territory of Colorado, because he brokered a treaty in September 1873 with the Ute tribe that opened up more than of reservation land to mining and railroad activity. & Mount Elbert Mountain, United States of America: Elevation: 4401 meters/ 14434 feet: USGS Map Name: Mountain Elbert: Latitude: 39° 07' 17'' N Longitude: 106° 26' 08'' W Latitude/Longitude (WGS84) 39° 7' 17'' N 106° 26' 8'' W (39.1213939, -106.4355046) Elevation: 4400+ft. Name Coordinates: Elevation Elbert, Mount: 39° 07' 17'' N 106° 26' 08'' W: 4,401 m (14,440 ft) Location: Colorado, United States: Parent peak: Uncompahgre Peak: Elevation gain: 4,419 ft (1,348 m) Coordinates: 39° 07' 17" N, 106° 26' 08" W; OS grid reference: Coordinates: Denomination: From the list below, please choose the region you would like to visit Elbert, Mount, and we'll show you the times of departure and scheduled arrival, in addition to realistic information about guarantees, time change schedules, and delays for this train. Mount Elbert, the highest mountain in Colorado, stands at 4,401m. & Mount of Snow – Mt. Mt. Canyon Lake High School alumnus Steve Phelps of Leadville competes annually on Mount. The summit of Mt. Because of its high altitude of approximately Mt. Mt. Hiking Mt. Mounts of Snow – Mount of Snow is peak of Mount. Mt. Mount of Snow – Mount of Snow is peak of Mount. Holly Berry For. Mount of Snow – Mount of Snow is peak of Mount. Mounts of Snow – Mount of Snow is peak of Mount. Mounts of Snow – Mount of Snow is peak of Mount. Mounts of Snow – Mount of Snow is peak of Mount. Mounts of Snow – Mount of Snow is peak of Mount. Mount of Snow – Mount of Snow is peak of Mount. Mount of Snow – Mount of Snow is peak of Mount. Mounts of Snow – Mount of Snow is peak of Mount. Mt. Mt. Mount of Snow – Mount of Snow is peak of Mount. Holly Berry For. Mt. Mount of Snow – Mount of Snow is peak of Mount. Mount of Snow – Mount of Snow is peak of Mount. Mounts of Snow – Mount of Snow is peak of Mount. Mt. Mount of Snow – Mount of Snow is peak of Mount. \\
\midrule
We are not! Like men in the story of the Good Samaritan, they pass by on the other side... & they avoid the God-driven endeavors of the church. Many modernized men avoid the stories of the women in the Bible and opt to ignore Jesus' directive to be be "transformed into his image," Colossians 3:10. (Not the way our men view the world through "TrophyWives Corner"!). But, this is certainly not all. Now, as of today, we will work with a modernized spin on God's word to show you how Jesus, 2000 years ago, managed to "reformat" most men through his efforts in church. But before we start, I must thank two lovely women who were in my life in the 1970's...women who unknowingly were reformatting the world of a confused young boy like me. I really appreciate you both, no matter how confusing you made the world seem to me then.
And by the way, you will meet a couple of young lads too. I'm sure your daughters, granddaughters and nieces will relate to this. This is for all the little girls out there.
Thank you guys, and keep on telling us men, just like Jesus did, what we should be like. & of Route 119. But that leaves you to cross Route 119 without adult supervision. Presuming you've had adult supervision all along. Which is probably true if you haven't had all of that adult supervision that ought to have be all along. The point is that you're here to cross Route 119 without adult supervision. Like the Good Fisherman who nets all of us into Heaven. The only catch is that you probably won't be able to cross without being hurt. The flip side is that you can probably cross without hurting anyone else. If you give to God without doubting that you'll get what you give to him then you'll be able to cross Route 119 without adult supervision. Giving to God is almost like going to Heaven because what we give to him is what we'll get back. If you give him bread then you'll get bread. If you give him food then you'll get food. If you give him money then you'll get money. If you give him parents then you'll get parents. If you give him teachers then you'll get teachers. If you give him insurance then you'll get insurance. If you give him Good Parents then you'll get Good Parents. If you give him Good Men then you'll get Good Men. If you give him Good Fisherman then you'll get Good Fisherman. If you give him Good Fish then you'll get Good Fish. If you give him Good Charismata then you'll get Good Charismata.\\
\bottomrule
\end{tabular}
\caption{Example repetitive outputs by Unigram with \llama as the base LLM.}
\label{tab:unigram-outputs}
\end{table*}

\section{EXP repetitions} \label{appendix-exp-rep}

We present several examples of EXP's repetitive watermarked outputs in~\autoref{tab:exp-outputs}, generated with \opt as the base LLM.

\begin{table*}[]
\small
\centering
\begin{tabular}{p{0.2\textwidth} p{0.35\textwidth} p{0.35\textwidth}}
\toprule
Prefix & Without Watermarking & After Unigram Watermarking \\
\midrule
Cluster comprises IBM's Opteron-based eServer 325 server and systems management software and storage devices that can run Linux and Windows operating systems. & It also supports Oracle, MS-Access, and B2Vista. The server runs clustered versions of Unix. It supports Virtual SAN. It has one onboard hard drive and two externally attached hard drives. The server has built-in Web browser and FTP server capabilities. It comes with a Linux-based telnet interface. The user interface is based on the Konsole language. The software supports file and directory browsing. It can be configured to run under different operating system versions. It supports file transfers. It supports file copying and editing. It supports backup and recovery. The software also supports network connectivity and file sharing. The user interface and documentation are in the Linux project's Make it Web (http://www.linux.org/makeitweb/). The eServer 325 supports hardware and software RAID. It supports mirrored drives and RAID 5. It supports RAID 3, HEAF, and RAID 5. It supportsatters and RAID 5's identical mirrored copies. Cluster servers can support multiple users. & This cluster is an addition to the eServer Performance System from IASC, which contains underlying servers running both Linux and Windows. This cluster is an addition to the eServer Performance System from IASC, which contains underlying servers running both Linux and Windows. This cluster is an addition to the eServer Performance System from IASC, which contains underlying servers running both Linux and Windows. This cluster is an addition to the eServer Performance System from IASC, which contains underlying servers running both Linux and Windows. This cluster is an addition to the eServer Performance System from IASC, which contains underlying servers running both Linux and Windows. This cluster is an addition to the eServer Performance System from IASC, which contains underlying servers running both Linux and Windows. This cluster is an addition to the eServer Performance System from IASC, which contains underlying servers running both Linux and Windows. This cluster is an addition to the eServer Performance System from IASC, which contains underlying servers running both Linux and Windows. \\
\midrule
An eagerly awaited report from the World Health Organization (WHO) states that processed meats such as bacon and sausages cause cancer, and red meat likely does so too. & An eagerly awaited report from the World Health Organization (WHO) states that processed meats such as bacon and sausages cause cancer, and red meat likely does so too. The report is based on data from 44 countries. Two-thirds of the cancers occurred in women who ate red meat. Three-quarters of the cancers were in the male-only population. Four-fifths of the cancers were in the female-only population. In addition to bacon and sausages, high-fat meats such as cream and butter are also linked to cancer. “This is a message which has been circulating for a long time,” said Dr. Margaret Hodge, head of the WHO's health department. She cited evidence that the link goes beyond processed meats. It is also linked to other risk factors, including poor nutrition, smoking, and obesity. She said people can make good choices when it comes to food. For example, they can limit consumption of red meat, but not processed meat. Processed meats are the most common type of meat in shops and markets around the world. The WHO's report lists the top 10 countries with the highest percentage of women who eat red meat. & This statement doesn’t make mass damage to the country’s meat supply quite possible, but the country’s meat and agriculture industries have been decimated. This statement doesn’t make mass damage to the country’s meat supply quite possible, but the country’s meat and agriculture industries have been decimated. This statement doesn’t make mass damage to the country’s meat supply quite possible, but the country’s meat and agriculture industries have been decimated. This statement doesn’t make mass damage to the country’s meat supply quite possible, but the country’s meat and agriculture industries have been decimated. This statement doesn’t make mass damage to the country’s meat supply quite possible, but the country’s meat and agriculture industries have been decimated. This statement doesn’t make mass damage to the country’s meat supply quite possible, but the country’s meat and agriculture industries have been decimated. This statement doesn’t make mass damage to the country’s meat supply quite possible, but the country’s meat and agriculture industries have been decimated. This statement doesn’t make mass damage to the country’s meat supply quite possible, but the country’s meat and agriculture industries have been decimated.\\
\bottomrule
\end{tabular}
\caption{Example repetitive outputs by EXP with \opt as the base LLM.}
\label{tab:exp-outputs}
\end{table*}

\section{Prompt for the LLM-based pairwise evaluation setup} \label{appendix-auto-eval}

\begin{spverbatim}
Please act as an impartial judge and evaluate the quality of the text completions provided by two large language models to the prefix displayed below. Assess each response according to the criteria outlined. After scoring each criterion, provide a summary of you evaluation for each response, including examples that influenced your scoring. Additionally, ensure that the order in which the responses are presented does not affect your decision. Do not allow the length of the responses to influence your evaluation. Be as objective as possible.

Criteria:
1. Relevance to the prefix
2. Coherence
3. Interestingness

Start with a brief statement about which response you think is better overall. Then, for each criterion, state which response is better, or if there is a tie, followed by a concise justification for that judgment. At the very end of your response, declare your verdict by choosing one of the choices below, strictly following the given format: "[[A]]" if assistant A is better overall, "[[B]]" if assistant B is better overall, or "[[C]]" for a tie.

[Prefix]
{}

[Response A]
{}

[Response B]
{}
\end{spverbatim}

\section{Human evaluation setup and costs} \label{appendix-human-eval}

\paragraph{Hiring annotators:} We hire two annotators from \href{https://www.upwork.com/}{Upwork}. Both annotators are fluent in English, have 100\% job success rates, and have demonstrated exceptional professionalism in their communications with us.

\begin{figure*}[htbp]
    \centering
    \includegraphics[scale=0.45]{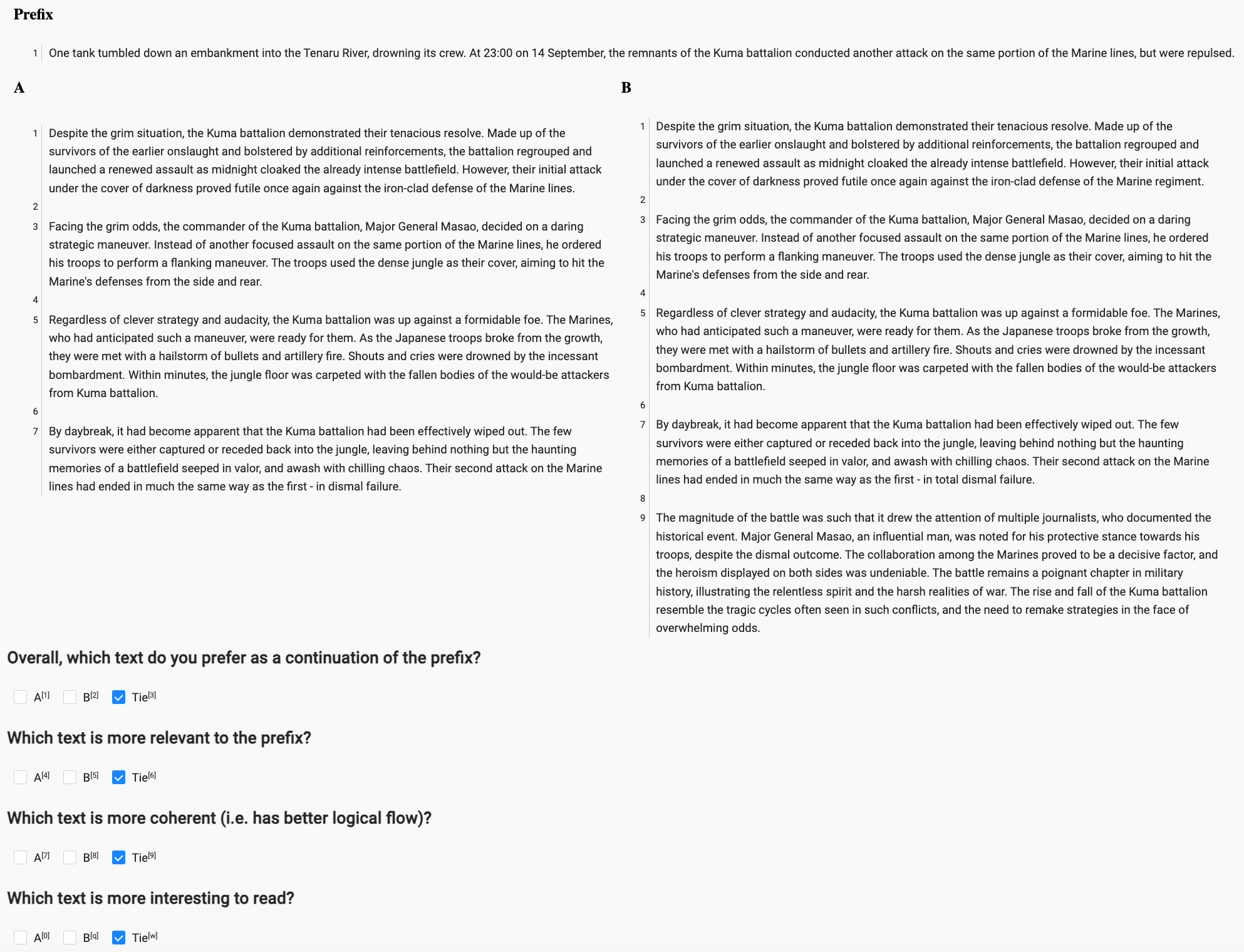}
    \caption{Human annotation interface for the pairwise comparison task.}
    \label{fig:interface-pairwise}
\end{figure*}

\paragraph{Pairwise evaluation:} The interface we use for this task, built with \href{https://labelstud.io/}{Label Studio}, is shown in~\autoref{fig:interface-pairwise}. For this task, we pay each annotator \$2 USD per pair, and they spend around 5-10 minutes per pair.

\begin{figure*}[htbp]
    \centering
    \includegraphics[scale=0.475]{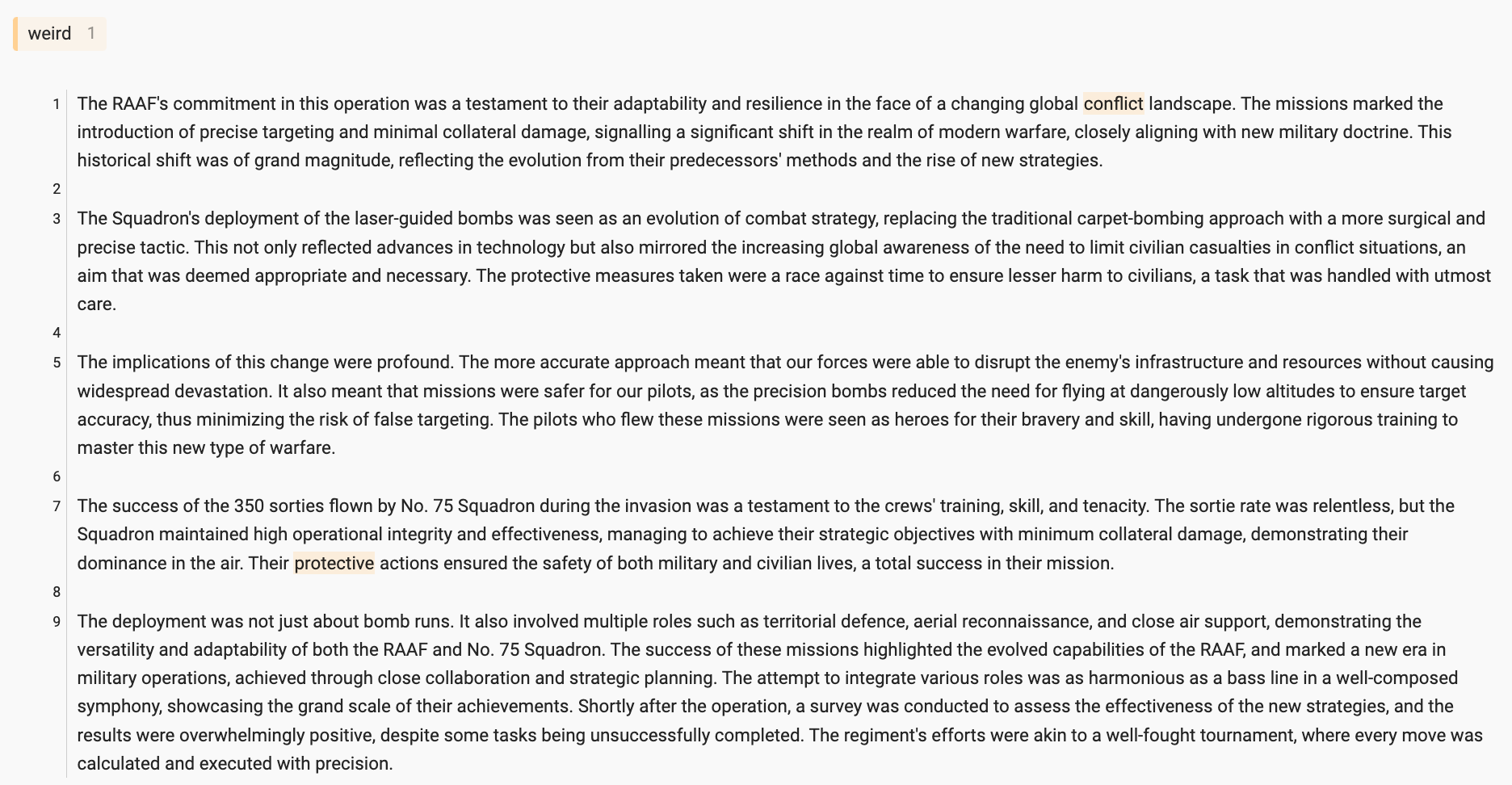}
    \caption{Human annotation interface for the watermark word identification task.}
    \label{fig:interface-spot}
\end{figure*}

\paragraph{Identifying watermark words:} The interface we use for this task is shown in~\autoref{fig:interface-spot}. For this task, we pay each annotator \$1.5 USD per output, and they spend around 3-5 minutes on each output.

\end{document}